\documentclass[preprint,11pt]{elsarticle}

\usepackage{amssymb}
\usepackage{booktabs}
\usepackage{amsthm}
\usepackage{mathtools}
\usepackage{algorithm}
\usepackage{algpseudocode}
\usepackage{caption}
\usepackage{subcaption}
\usepackage[top=2.5cm,bottom=2.0cm,left=2.0cm,right=2.5cm]{geometry}
\usepackage{enumitem} 
\usepackage[flushleft]{threeparttable}
\usepackage{lineno,hyperref}
\journal{Structural Safety}

\begin{document}

\begin{frontmatter}


\title{Optimal Inspection and Maintenance Planning for Deteriorating Structural Components through Dynamic Bayesian Networks and Markov Decision Processes}
%%\author[1]{P.G. Morato\corref{cor1}}
%%\ead{pgmorato@uliege.be}
%% \cortext[cor1]{Corresponding author}
%% \address[1]{Address\fnref{label3}}

%%\title{Optimal Inspection and Maintenance Planning for Deteriorating Structures through Dynamic Bayesian Networks and Markov Decision Processes}


\author[1]{P.G. Morato\corref{cor1}}
\ead{pgmorato@uliege.be}
\author[2]{K.G. Papakonstantinou}
\author[2]{C.P. Andriotis}
\author[3]{J.S. Nielsen}
\author[1]{P. Rigo}

\cortext[cor1]{Corresponding author}
\address[1]{ANAST, Department of ArGEnCo, University of Liege, 4000, Liege, Belgium}
\address[2]{Department of Civil \& Environmental Engineering, The Pennsylvania State University, University Park, PA 16802, USA}
\address[3]{Department of the Built Environment, Aalborg University, 9220, Aalborg, Denmark}

\begin{abstract}
Civil and maritime engineering systems, among others, from bridges to offshore platforms and wind turbines, must be efficiently managed, as they are exposed to deterioration mechanisms throughout their operational life, such as fatigue and/or corrosion. Identifying optimal inspection and maintenance policies demands the solution of a complex sequential decision-making problem under uncertainty, with the main objective of efficiently controlling the risk associated with structural failures. Addressing this complexity, risk-based inspection planning methodologies, supported often by dynamic Bayesian networks, evaluate a set of pre-defined heuristic decision rules to reasonably simplify the decision problem. However, the resulting policies may be compromised by the limited space considered in the definition of the decision rules. Avoiding this limitation, Partially Observable Markov Decision Processes (POMDPs) provide a principled mathematical methodology for stochastic optimal control under uncertain action outcomes and observations, in which the optimal actions are prescribed as a function of the entire, dynamically updated, state probability distribution. In this paper, we combine dynamic Bayesian networks with POMDPs in a joint framework for optimal inspection and maintenance planning, and we provide the relevant formulation for developing both infinite and finite horizon POMDPs in a structural reliability context. The proposed methodology is implemented and tested for the case of a structural component subject to fatigue deterioration, demonstrating the capability of state-of-the-art point-based POMDP solvers of solving the underlying planning stochastic optimization problem. Within the numerical experiments, POMDP and heuristic-based policies are thoroughly compared, and results showcase that POMDPs achieve substantially lower costs as compared to their counterparts, even for traditional problem settings.
\end{abstract}


%\item A methodology for optimal inspection and maintenance planning is proposed.
%\item DBNs and POMDPs are combined in a joint framework.
%\item The formulation for deriving POMDPs in a structural reliability context is presented.
%\item POMDP and heuristic-based policies are thoroughly compared.
%\item POMDPs offer substantially lower costs as compared to their counterparts. 


\end{frontmatter}

%\linenumbers

\section{Introduction}\label{Sec:intro}
Preserving infrastructures in a good condition, despite their exposure to diverse deterioration mechanisms throughout their operational life, enables, in most countries, a stable economic growth and societal development \cite{Frangopol2011}.  For instance, a bridge structural component may experience a thickness reduction due to corrosion effects\citep{Corrosion_Stewart,corrosionBridgeFrangopol,CorrosionMelchersStrewart,CorrosionMelchers_1}; or a surface crack at an offshore platform might drastically propagate due to fatigue deterioration \citep{Moan_Offshore2005,MOAN_Fatigue_deterior,WirschingFatigueDeterior}; or the structural resistance of an offshore welded joint can be reduced due to the combined cyclic actions of wind and ocean waves \citep{SChaumanOWSupportStructures,Dong2012,FatigueSoares2015}. The prediction of such deterioration processes involves a probabilistic analysis in which all relevant uncertainties are properly quantified. 

Information about the condition of structural components can be gathered during their operational life through inspections or monitoring, allowing the decision maker to take more informed and rational actions \citep{andriotis2019value,fauriat2020optimization}. However, both maintenance actions and observations are associated with certain costs which must be optimally balanced against the risk of structural failure. As suggested by \citep{Elingwood2005, sanchez2016maintenance} and others, inspections and/or maintenance actions should be planned with the objective of optimizing the structural life-cycle cost. Besides economic consequences associated with structural failures or maintenance interventions, societal and environmental aspects can also be included within a decision-making context in terms of utilities, defined in monetary units. A decision maker should, therefore, identify the decisions that result in the minimization of the total expected costs over the lifetime of the structure \citep{Faber2003RiskIntroduction,Frangopol_FatiPLAN_2019}. 

\begin{figure}
	\centering
		\includegraphics[scale=1]{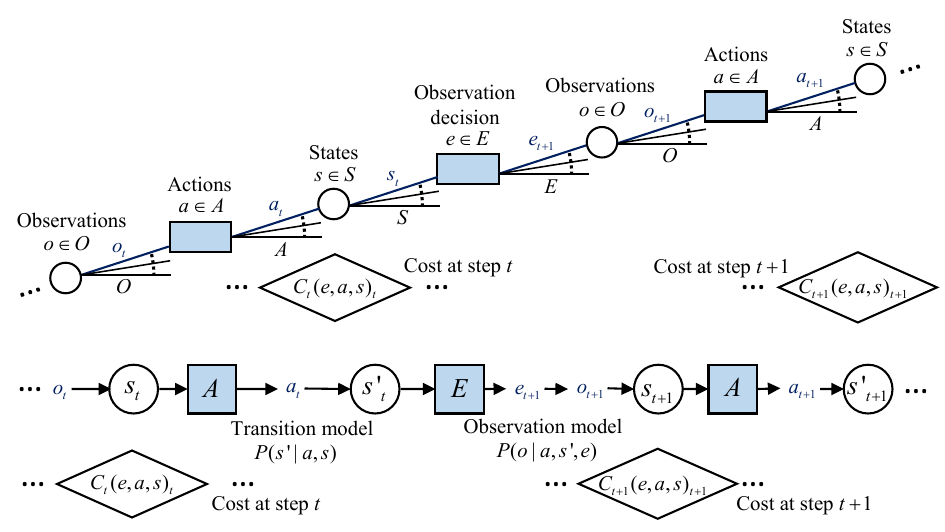}
	\caption{(Top) Inspection and Maintenance (I\&M) planning decision tree. Maintenance actions and observation decisions are represented by blue boxes and chance nodes are depicted by white circles. At every time step, the cost $C_t$ depends on the action $a$, observation decision $e$, and state $s$ of the component.  (Bottom) An I\&M POMDP sequence is represented where at each step $t$, the cost $C_t$ depends on the action $a$, observation decision $e$, and state $s$ of the component. In both representations, an observation outcome $o$ is collected according to the current state, taken action and observation decision.}
	\label{FIG:dectree}
\end{figure}

In the context of Inspection and Maintenance (I\&M) planning, the decision maker faces a complex sequential decision-making problem under uncertainty. This sequential decision-making problem is illustrated in Figure \ref{FIG:dectree}, showcasing the involved random events and decision points, and can be formulated either from the perspective of the classical applied statistical decision theory \citep{Raiffa1961}, or through artificial intelligence \citep{AI_bookRussell} conceptions, or a combination thereof. In all cases, the main objective of a decision maker, or an intelligent agent, is to identify the optimal policy that minimizes the total expected costs.

With the aim of addressing this complex decision-making problem, Risk-Based Inspection (RBI) planning methodologies have been traditionally proposed \citep{Soren_EarlyRBI} and have often also been applied to the I\&M planning of offshore structures \citep{RBI_Faber_Goyet2002,Rangel-Ramirez2012}. By imposing a set of heuristic decision rules, RBI methodologies are able to simplify and solve the decision-making problem within a reasonable computational time, while structural reliability methods are often employed within this framework, to quantify and update the reliability and risk metrics.

More recently, RBI methodologies have also been integrated with Dynamic Bayesian Networks (DBNs) \citep{StraubDBN2009,LuqueDBN2019,bismut2021optimal,yang2018DBNprobabilistic,tien2017reliability}. DBNs provide an intuitive and robust inference approach to Bayesian updating; however, they do not tackle the decision optimization problem by themselves. In the proposed methodologies, heuristic decision rules, usually based on engineering principles and understanding of the problem, are still utilized to simplify the decision problem. Albeit their practical advantages, the main shortcoming of heuristic-based policies is the limited policy space exploration due to the prior, ad-hoc prescription of decision rules. In this paper, we thus present how DBNs describing deterioration processes can be instead combined with Markov decision processes and dynamic programming \citep{puterman2014markov}, and be used to define transition and emission probabilities in such settings.

Partially Observable Markov Decision Processes (POMDPs) provide a principled mathematical methodology for planning in stochastic environments under partial observability. In the past, POMDPs were only applicable for small state space problems due to the difficulty of finding appropriate solutions in a reasonable computation time. However, starting with the development of point-based solvers \citep{shani2013survey}, which managed to efficiently alleviate the inherent complexities of the solution process, POMDPs have been increasingly used for planning problems, especially, in the field of computer science and robot navigation \citep{sarsop_POMDP,FRTDP_Conf2006}. POMDPs have also been proposed for I\&M of engineering systems \citep{POzziMilad2015,Pozzi_Hierarchical_2016,Papakonstantinou2014Part1,Corotis_POMDP_2005,POMDP_Pablo_ICASP}. 
In the reported POMDP methodologies, either the condition of the structural component has been modeled with less than five discrete states or the rewards have not been defined in a structural reliability context. This different POMDP approach to the I\&M problem, as compared with typical RBI applications, has raised some misconceptions in the literature about their use, which we formally rectify herein.

In this work, POMDPs are successfully combined with dynamic Bayesian networks in a joint framework, for optimal inspection and maintenance planning, in order to take advantage of both the modeling flexibility of DBNs and the advanced optimization capabilities of POMDPs. In particular, this paper originally derives the POMDP dynamics from DBNs, enabling optimal control of physically-based stochastic deterioration processes, modeled either through a conditional set of time-invariant parameters or as a function of the deterioration rate. We further provide all relevant formulations for deriving both infinite and finite horizon POMDPs within a structural reliability context. The proposed framework is analyzed, implemented, and tested for the case of a structural component subject to a fatigue deterioration process, and the capability of state-of-the-art point-based POMDP value iteration methods to efficiently solve challenging I\&M optimization problems is verified. POMDP and typical heuristic risk-based and/or periodic policies are thoroughly analyzed and compared, in a variety of problem settings, and results demonstrate that POMDP solutions achieve substantially lower costs in all cases, as compared to their counterparts.

\section{Background: Risk-based inspection planning}\label{Sec:RBI}
A typical Inspection and Maintenance (I\&M) sequential decision problem under uncertainty is illustrated in Figure \ref{FIG:dectree}. The optimal strategy can be theoretically identified by means of a pre-posterior decision analysis \citep{Raiffa1961}. Assuming the costs at different times to be additive independent, the pre-posterior analysis prescribes the observation decisions $e\in E$ and actions $a\in A$ that minimize the total expected cost $C_T(a,e)= C_{t_0}(e,a,s)_{t_0} + ... + C_{t_N}(e,a,s)_{t_N}\gamma^{t_{N}}$, i.e. the sum over the lifetime $t_N$ of the discounted costs received at each time step $t$, with $\gamma$ being the discount factor. Note that societal and environmental consequences, specified in monetary units, can also be included within the definition of the total expected cost.

If the probabilities associated with the random events, as well as the costs, are assigned to each branch of the decision tree, then the branch corresponding to the optimal cost $C_{T}^*(a,e)$ can be identified. This analysis is denoted backwards induction or extensive analysis. Alternatively, a normal analysis can also be conducted by identifying the optimal decision rule, $h^*_{a,e}$, from all possible decision rules. In any case, the exact solution of a pre-posterior analysis very quickly becomes computationally intractable for practical problems because the number of branches increases exponentially with the number of time steps, actions, and observations. 

\subsection{RBI assumptions and heuristic rules}\label{Subsec:heurules}
Risk-Based Inspection (RBI) planning methodologies \citep{Faber2002RBI} introduce simplifications to the I\&M decision-making problem in order to be able to identify strategies in a reasonable computational time. To simplify the problem, the expected cost is computed only for a limited set of pre-defined decision rules $h_{a,e}$. The best strategy among them is then identified as the decision rule with the minimum cost.

Within an I\&M planning context, the total expected cost $\mathbf{E}[C_T(h,t_N)]$ is the combination of expected costs from inspections $\mathbf{E}[C_{I}(h,t_{N})]$, repairs $\mathbf{E}[C_{R}(h,t_{N})]$, and failures $\mathbf{E}[C_{F}(h,t_{N})]$, as a function of the imposed decision rules $h_{a,e}$. This expectation for a structural component designed for a lifetime of $t_N$ years is simply computed as:
\begin{equation}\label{Eq:Util1}
\mathbf{E}[C_T(h,t_N)]=\mathbf{E}[C_{I}(h,t_{N})] +\mathbf{E}[C_{R}(h,t_{N})]+\mathbf{E}[C_{F}(h,t_{N})]
\end{equation}
The simplifications introduced to the I\&M decision-making problem by pre-defining a set of decision rules are listed below:
\begin{enumerate}[label=\roman*)]
\item Observations (inspections) are planned according to a pre-defined heuristic rule. Two heuristic rules are commonly employed in the literature  \citep{Straub2004Thesis}:
\begin{itemize} 
\item Equidistant inspections: inspections are planned at constant intervals of time $\Delta t$. 
\item Failure probability threshold: inspections are planned just before a pre-defined annual failure probability $\Delta P_F$ threshold is reached.
\end{itemize}
\item If the outcome of an inspection indicates damage detection $(d>d_{det})$, a repair action is immediately performed. In that case, the repair probability is equal to the probability of detection $P_R=P(d>d_{det})$. Alternatively, other heuristic rules can also be imposed (adding computational complexity), such as that a repair is performed if an inspection indicates detection $(d>d_{det})$ and a pre-defined failure probability threshold $P_F$ is simultaneously exceeded.
\item After a component is repaired, it is assumed that it behaves like a component with no damage detection, i.e. the remaining life can be computed as if the inspection at the time of repair indicates no damage detection. With these assumptions, the decision tree represented in Figure \ref{FIG:dectree} can be simplified to a single branch. Alternatively, if a repair is performed at time $t$ and it is assumed to be perfect, the component returns to its initial damage state at the beginning of a new decision tree with a lifetime equal to $\bar{t}_N=t_N-t$. 
\end{enumerate}

Summarizing, one can simplify the problem to one decision tree branch by assuming that: (i) inspections are to be planned according to a heuristic rule, (ii) a repair is to be performed if an inspection indicates detection, and (iii) after a repair is performed, the inspection at that time is considered as a no detection event. In this case, the individual contributions to the total expected cost in Equation \ref{Eq:Util1} can be computed analytically. 
\vfill
\noindent The expected inspection cost $\mathbf{E}[C_I(h,t_{N})]$ is computed as the sum of all conducted inspections $I_n$, with individual inspection cost $C_i$, and discounted by the factor $\gamma \in [0,1]$:
\begin{equation}\label{Eq:RBICI}
\mathbf{E}[C_I(h,t_{N})]=\sum_{t_{I}=t_{I_1}}^{t_{I_n}}C_{i}\gamma^{t_I}
\end{equation}
The expected repair cost $\mathbf{E}[C_R(h,t_{N})]$ corresponding to a heuristic scheme $h$ is calculated as the repair cost $C_r$ multiplied by the probability of repair $P_R$ at each inspection year $t_I$:
\begin{equation}\label{Eq:RBICR}
\mathbf{E}[C_R(h,t_{N})]=\sum_{t_{I}=t_{I_1}}^{t_{I_n}}C_{r}P_R(h,t)\gamma^{t_I}
\end{equation}
The expected risk of failure $\mathbf{E}[C_F(h,t_{N})]$ is computed as the sum of discounted annual failure risks, in which $\Delta P_F$ is the annual failure probability and $C_f$ is the cost of failure:
\begin{equation}\label{Eq:RBICF}
\mathbf{E}[C_F(h,t_{N})]=\sum_{t=1}^{t_{N}}C_{f}\Delta P_F(h,t)\gamma^{t}
\end{equation}

\subsection{Probabilistic deterioration model and reliability updating}\label{Subsec:RBIUpd}

Structural reliability methods and general sampling based methods \citep{Ditlevsen2007StructuralMethods} can be used to compute the probabilities associated with the random events represented in the I\&M decision tree (Figure \ref{FIG:dectree}). In a simplified decision tree, the main random events are the damage detections during inspections and the structural failure.

The failure event is defined through a limit state $g_F(t) = d_c - d(t)$, in which $d_c$ represents the failure criteria, such as the critical crack size, and $d(t)$ is related to the temporal deterioration evolution. Uncertainties involved in the deterioration process are incorporated by defining $d(t)$ as a function of a group of random variables or random processes. The probability of failure $P_F(t)$ can be then computed as the probability of the limit state being negative $P_F=P\{g_F(t)\leq0\}$, and the reliability index is inversely related to the failure probability, usually defined in the standard normal space as $\beta(t)=-\Phi^{-1}\{P_F(t)\}$, in which $\Phi$ is the standard normal cumulative distribution function. The probability of the failure event can also be defined over a reference period, e.g. the annual failure probability can be computed as $\Delta P_F(t) = \{P_F(t) - P_F(t-1)\}$.

The measurement uncertainty of the available observations (inspections) is often quantified by means of Probability of Detection (PoD) curves. A PoD indicates the probability of detection as a function of the damage size $d$ and depends on the employed inspection method, i.e. the function of the detectable damage size can be modeled by an exponential distribution $F(d_d)=F_0\big[1-exp(-d/\lambda)\big]$, where $F_0$ and $\lambda$ are parameters determined by experiments. The event of no detection at time $t_I$ is then modeled by the limit state function $g_{I_{nd}}(t_I)=d(t_I)-d_d(t_I)$. Similarly, the event of detection at time $t_I$ is modeled by the limit state $g_{I_d}(t_I)=d_d(t_I)-d(t_I)$. Both detection and no detection events are evaluated as inequalities, for instance, the probability of no detection is assessed as the probability of the limit state being negative $P_{I_{nd}}=P\{g_{I_{nd}}(t_I)\leq0\}$. Alternatively, a discrete damage measurement $d_m$ can be collected and the limit state is modeled in this case as $g_m(t_I)=d(t_I)-(d_m-\epsilon_m)$, where $\epsilon_m$ is a random variable that represents the measurement uncertainty, and the equality event $P_m=P\{g_m(t)=0\}$ can be estimated equal to some limit, as explained in \citep{Ditlevsen2007StructuralMethods,madsen2006methods,straub2011EQreliability}. 
\clearpage
The additional information gained by observations can be used to update the structural reliability or failure probability $P_F$ by computing a failure event conditional on inspection events \citep{Lotsberg2016ProbabilisticStructures}, as:

\begin{equation} \label{Eq:Upd1}
P_{F|I_1,...,I_N}(t)=\frac{P\big[g_F(t)\leq0\cap g_{I_1}(t)\leq0\cap ...\cap g_{I_N}(t)\leq0\big]}{P\big[g_{I_1}(t)\leq0\cap ...\cap g_{I_N}(t)\leq0\big]}
\end{equation} 
The conditional failure probability introduced in Equation \ref{Eq:Upd1} can be computed by structural reliability methods (FORM, SORM) or by Monte Carlo sampling methods \citep{Ditlevsen2007StructuralMethods}.  

\section{Stochastic deterioration processes via Dynamic Bayesian Networks}\label{Sec:DBN}
A brief overview on the adoption of dynamic Bayesian networks (DBNs) for structural deterioration and reliability problems is presented here, with the objective of demonstrating that the main principles underlying BNs inference tasks are fundamentally invariant to those employed by POMDPs. Bayesian networks (BNs) are directed acyclic graphical models particularly suited for inference tasks in probabilistic environments. A DBN is a template model of a Bayesian network evolving over time and in the context of structural reliability and related problems, DBNs have played an important role \citep{StraubDBN2009,LuqueDBN2019,DBNLuque2016}.  For a detailed background of probabilistic graphical models and BNs, the reader is directed to \citep{jensenbookDBN}. 

To allow DBNs based inference within a reasonable computational time for practical problems, the following assumptions are often imposed:

\begin{enumerate}[label=\roman*)]
\item Discrete state space: Exact inference algorithms are limited to discrete random variables \citep{murphythesis}. A discretization operation must thus be performed to convert the original continuous random variables to the discrete space. The unknown error introduced by the discretization operation converges to zero in the limit of an infinitesimal interval size. However, the computational complexity of the inference task grows linearly with the number of states and exponentially with the number of random variables.
\item Markovian assumption: The state space $S$ is the domain of all random variables involved in the description of the deterioration process, and the conditional probabilities $P(s_{t+1}|s_t)$ associated with the random variables at time step $t+1$ depend only on the random variables at the current time step $t$, and are independent of all past states.
\end{enumerate}
The transition probability matrix $P(s_{t+1}|s_{t})$ can also be assumed as stationary for some applications, thus facilitating the formulation of the problem. This can however be easily relaxed without entailing additional computational efforts \citep{Jannie2018Computational}.

\subsection{Parametric DBN}\label{Subsec:DBNparam}
\begin{figure}
	\centering
		\includegraphics[scale=1]{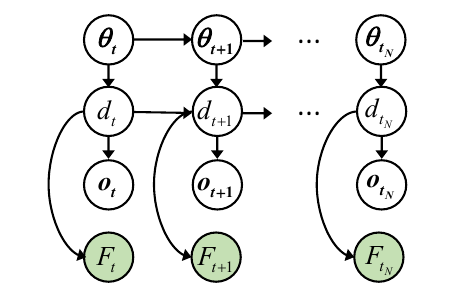}
	\caption{Parametric dynamic Bayesian network, adapted from \citep{StraubDBN2009}. The evolution of a stochastic deterioration process is represented by the nodes $d_t$ influenced by a set of time-invariant random variables $\boldsymbol{\theta_t}$. Imperfect observations are added through the nodes $\boldsymbol{o_t}$, and $F_t$ binary node indicates the probability of failure and survival events.}
	\label{FIG:dbngraph}
\end{figure}
A stochastic deterioration process can be represented by the DBN shown in Figure \ref{FIG:dbngraph}. The deterioration is represented through the damage node $d_t$ which is influenced by a set of time-invariant random variables $\boldsymbol{\theta_t}$. The model is denoted as parametric DBN as the damage $d_t$ is influenced by the parameters $\boldsymbol{\theta_t}$.  Imperfect observations are added into the DBNs by means of the node $\boldsymbol{o_t}$. This DBN can be extended by incorporating time-variant random variables as proposed by \citep{StraubDBN2009}; yet, we consider only time-invariant random variables here as they are widely used in the literature and to avoid unnecessary presentation complications. Finally, the binary node $F_t$ provides an indication of the failure and survivability.

Within the context of structural reliability and related problems, DBNs are often employed to propagate and update the uncertainty related to a deterioration process, incorporating evidence from inspections or monitoring. Filtering becomes the preferred inference task for inspection and maintenance planning problems, as a decision is taken at time $t$ supported by evidence gathered from the initial time step $t_0$ up to time $t$. The belief state, defined as the probability distribution over states, can be propagated and updated by applying the forward operation from the forward-backward algorithm \citep{murphythesis}. The transition algorithmic step of the forward operation is assumed to be Markovian, being therefore equivalent to the underlying transition model of a POMDP. More details on the formulation of POMDP transition models are introduced in Section \ref{Sec:POMDPImplem}.  

At time step $t_0$, the initial belief corresponds to the joint probability of the initial damage and time-invariant parameters $P(d_{t_0},\boldsymbol{\theta_{t_0}})$. The forward operation is then applied for the subsequent time steps, comprised of the following steps:
\begin{enumerate}
  \item Transition step: the belief propagates in time according to a pre-defined conditional probability distribution or transition matrix $P(d_{t+1},\boldsymbol{\theta_{t+1}}|d_{t},\boldsymbol{\theta_{t}})$, as:

\begin{equation} \label{Eq:forwOp1}
P(d_{t+1},\boldsymbol{\theta_{t+1}}|\boldsymbol{o_0},...,\boldsymbol{o_{t}}) = \sum_{d_{t}} \sum_{\boldsymbol{\theta_{t}}} P(d_{t+1},\boldsymbol{\theta_{t+1}}|d_{t},\boldsymbol{\theta_{t}}) \, P(d_{t},\boldsymbol{\theta_{t}}|\boldsymbol{o_0},...,\boldsymbol{o_{t}})   
\end{equation}
  \item Estimation step: the belief is now updated based on the obtained evidence by means of Bayes' rule, as: 
\begin{equation} \label{Eq:forwOp2}
P(d_{t+1},\boldsymbol{\theta_{t+1}}|\boldsymbol{o_0},...,\boldsymbol{o_{t+1}}) \propto P(\boldsymbol{o_{t+1}}|d_{t+1})  
P(d_{t+1},\boldsymbol{\theta_{t+1}}|\boldsymbol{o_0},...,\boldsymbol{o_{t}})
\end{equation}
The quality of the observation is quantified by the likelihood $P(\boldsymbol{o_{t+1}}|d_{t+1})$. This likelihood can be directly obtained from probability of detection curves or by discretizing a direct measurement. Since the random variables are discrete, a normalization of $P(d_{t+1}, \boldsymbol{\theta_{t+1}}|\boldsymbol{o_0},...,\boldsymbol{o_{t+1}})$ can be easily implemented.
\end{enumerate}

The failure probability assigned to the node $F_t$ corresponds to the probability of being in a failure state. As the failure states are defined based on the damage condition $d_t$, the time invariant parameters $\boldsymbol{\theta_{t}}$ can be marginalized out to compute the failure probability. Disregarding the discretization error, the resulting structural reliability is equivalent to the one computed in Equation  \ref{Eq:Upd1}. 

In terms of computational complexity, note that the belief is composed of $(|\theta_1|\cdot ... \cdot |\theta_k| \:|d|)$ states, defined by the damage $d$ along with $k$ time-invariant random variables. Thus, the transition matrix includes $(|\theta_1|\cdot ... \cdot |\theta_k| \:|d|)^2$ elements. Since $P(\theta_{t+1}|\theta_{t})$ is defined by an identity matrix, the transition is prescribed by a very sparse, block-diagonal matrix with a maximum density of $\rho_{P}=1/(|\theta_1|\cdot ... \cdot |\theta_k|)$.
\clearpage

\subsection{Deterioration rate DBN}\label{Subsec:DBNdetrat}
\begin{figure}
	\centering
		\includegraphics[scale=1]{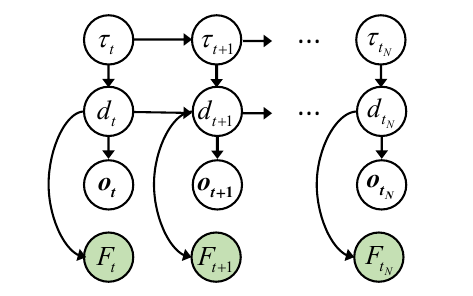}
	\caption{Deterioration rate dynamic Bayesian network, derived from \citep{Papakonstantinou2014a}. The evolution of a stochastic deterioration process is represented by the nodes $d_t$ dependent on the deterioration rate ${\tau_t}$. Imperfect observations are included through the nodes $\boldsymbol{o_t}$, and $F_t$ binary node indicates the probability of failure and survival events.}
	\label{FIG:dbngraphDR}
\end{figure}

We present herein an alternative DBN in which a stochastic deterioration process is represented as a function of the deterioration rate. This model is adopted from \citep{Papakonstantinou2014a} and denoted here as deterioration rate DBN. Figure \ref{FIG:dbngraphDR} graphically illustrates the model. In this case, the stochastic deterioration process is described in time $t$ by the nodes $d_t$, conditional on the deterioration rate $\tau_t$. If the stochastic process is stationary, the deterioration evolution will vary equally over time, and thus the deterioration rate $\tau_t$ is not utilized. The deterioration does not, however, progress equally over time in a non-stationary process, and in that case, the parameter $\tau_t$ needs to be incorporated to effectively model the varying deterioration effects over time. After collecting experimental or physically-based simulated data (e.g. Monte Carlo simulations) from a non-stationary deterioration process, the transition probabilities can be calculated, for each deterioration rate $\tau_t$, by counting the number of transitions from $d_{t}$ to $d_{t+1}$ over the total data available in $d_t$. Additional methods to compute the transition model are described in \citep{Papakonstantinou2014a}. As illustrated in Figure \ref{FIG:dbngraphDR}, imperfect observations are added through the nodes $\boldsymbol{o_t}$ and the structural reliability is indicated through the node $F_t$.

To ensure compliance with the DBNs time invariant property, the belief incorporates both the damage condition and deterioration rate through the joint probability $P(d_{t},\tau_{t})$. Yet, the node $\tau_{t}$ is a zero-one vector (one-hot) that transitions each time step from one deterioration rate $\tau_i$ to the next $\tau_{i+1}$. The deterioration evolution is computed by a forward operation in a similar manner as for the parametric DBN. Initially, the belief corresponds to the joint probability $P(d_{0},\tau_{0})$. Subsequently, the belief experiences a transition according to the transition matrix $P(d_{t+1},\tau_{t+1}|d_{t},\tau_{t})$:

\begin{equation} \label{Eq:forwOpDR}
P(d_{t+1},\tau_{t+1}|\boldsymbol{o_0},...,\boldsymbol{o_{t}}) = \sum_{d_{t}} \sum_{\tau_{t}} P(d_{t+1},\tau_{t+1}|d_{t},\tau_{t}) \, P(d_{t},\tau_{t}|\boldsymbol{o_0},...,\boldsymbol{o_{t}}) 
\end{equation}

Based on the gathered observations, the beliefs are then updated by applying Bayes' rule. The likelihood $P(\boldsymbol{o_{t+1}}|d_{t+1})$ can be directly defined from probability of detection curves or other observation uncertainty measures:

\begin{equation} \label{Eq:forwOpDR2}
P(d_{t+1},\tau_{t+1}|\boldsymbol{o_0},...,\boldsymbol{o_{t+1}}) \propto P(\boldsymbol{o_{t+1}}|d_{t+1})
 P(d_{t+1},\tau_{t+1}|\boldsymbol{o_0},...,\boldsymbol{o_{t}})
\end{equation}

The computational complexity is influenced by the belief size. For the case of a deterioration rate DBN, the belief $P(d_{t},\tau_{t})$ is composed of $|\tau|\cdot|d|$ states and its sparse transition matrix $P(d_{t+1},\tau_{t+1}|d_{t},\tau_{t})$ accounts for $(|\tau|\:|d|)^2$ elements. Since the only non-zero probabilities of the transition matrix $P(\tau_{t+1}|\tau_{t})$ are the ones to define the transition from deterioration rate $\tau_{t}$ to the next deterioration rate $\tau_{t+1}$, the maximum density of $P(d_{t+1},\tau_{t+1}|d_{t},\tau_{t})$ is $\rho_{DR}=1/|\tau|$. 

Advantages between a parametric DBN and a deterioration rate one are case dependent. If the deterioration process can be modeled by just few parameters or it evolves over a long time span, the parametric DBN is recommended. However, if the deterioration modeling involves many parameters or complex random processes spanning over a short time horizon, the deterioration rate DBN should be preferred. If both DBN models are applied for the same problem, the results should be equivalent and differences are only affected by the discretization error.

\subsection*{Risk-based inspection planning and DBNs}

While DBNs can be successfully used for reliability updating, they do not possess by themselves intrinsic optimization capabilities. To this end, modern RBI methodologies include a combination of DBNs and heuristic rules to identify the optimal strategy \citep{DBNLuque2016, Straub2009}. The methodologies often follow a similar logic as the theoretical scheme presented in Section \ref{Sec:RBI}, where the decision tree is simplified.

Alternatively, the optimal I\&M strategy among different alternatives can be identified with the support of DBNs in a simulation environment. Any of the proposed DBN types (Sections \ref{Subsec:DBNparam} and \ref{Subsec:DBNdetrat}) can be generalized to an influence diagram by adding utility and decision nodes \citep{LuqueDBN2019}. The total cost $C_T$ for a set of pre-defined heuristic rules $h_{a,e}$ can be computed by simulating one episode $ep$ of length $t_N$ as: 
\begin{equation} \label{Eq:DBNSim1}
C_{T_{ep}}(h) = \sum^{t_{N}}_{t=t_0} \Big[C_i(t)\gamma^t + C_r(t)\gamma^t + \Delta P_F(t)C_f\gamma^t\Big]
\end{equation}
The total expected cost $\mathbf{E}[C_{T}(h)]$ is then computed with a Monte Carlo simulation of $n_{ep}$ episodes (policy realizations): 
\begin{equation} \label{Eq:DBNSim2}
\mathbf{E}[C_{T}(h)] = \frac{\sum^{n_{ep}}_{ep=1}\big[C_{T_{ep}}(h)\big]}{n_{ep}}
\end{equation}
One can compute the costs of all pre-defined heuristic rules and identify the strategy with the minimum expected cost as the optimal policy. However, the resulting optimal policies might be compromised due to the limited space covered by the imposed heuristic rules, out of all possible decision rules.

\section{Optimal I\&M planning through POMDPs}\label{Sec:POMDP}
We propose herein a methodology for optimal I\&M planning of deteriorating structures under uncertainty based on Partially Observable Markov Decision Processes (POMDPs). The methodology is adopted by similar frameworks, as studied in \citep{Papakonstantinou2014Part2}. While the damage evolution was modeled in \citep{Papakonstantinou2014Part2} as function of its deterioration rate, following the formulation presented in Section \ref{Subsec:DBNdetrat}, we extend here the methodology to deterioration mechanisms modeled as functions of time-invariant parameters, formulated according to Section \ref{Subsec:DBNparam}. In addition, the user penalty is defined in this work as a consequence of the annual failure probability experienced by the component. 

A Markov decision process (MDP) is a 5-tuple $\langle S, A, T, R, \gamma \rangle$ controlled stochastic process in which an intelligent agent acts in a stochastic environment. The agent observes the component at state $s\in S$ and takes an action $a\in A$, then the state randomly transitions to state $s'\in S$ according to a transition probability model $T(s,a,s')=P(s'|s,a)$, and finally the agent receives a relevant reward $R_t(s,a)$, where $t$ is the current decision step. 

As described in Section \ref{Sec:intro}, the optimal decisions result in a minimum expected cost. The expected cost, or value function, is expressed for a finite horizon MDP as the summation of the decomposed rewards $V(s_0) = R_{t_0} + ... + R_{t_{N-1}}\gamma^{t_{N-1}}$, from time step $t_0$ up to the final time step $t_{N-1}$. For an infinite or unbounded horizon MDP, the rewards are infinitely summed up ($t_N=\infty$). Note that the rewards are discounted by the factor $\gamma$. From an economic perspective, the discount factor converts future rewards into their present value. Computationally, discounting is also necessary to guarantee convergence in infinite horizon problems.

An MDP policy ($\pi:S\rightarrow A$) prescribes an action as a function of the current state. The main goal of an MDP is the identification of the optimal policy $\pi^*(s)$ which maximizes the value function $V^*(s)$. There exist efficient algorithms that compute the optimal policy using the principles of dynamic programming and invoking Bellman's equation. Both value and policy iteration algorithms can be implemented to identify the optimal policy $\pi^*(s)$ \citep{DecisionMakBook_Kochenderfer}. While the state of the component in an MDP is known at each time step, imperfect observations are usually obtained in real situations, e.g. noise in the sensor of a robot, measurement uncertainty of an inspection, etc. 
\begin{figure}
	\centering
		\includegraphics[scale=1]{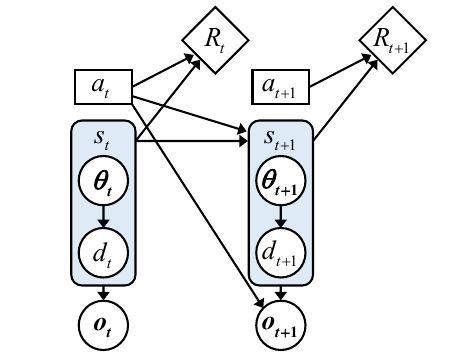}
	\caption{Graphical representation of a Partially Observable Markov Decision Process (POMDP). The states $S_t$ are modeled as the joint distribution of the time-invariant parameters $\boldsymbol{\theta_{t}}$ and the damage size $d_t$. The imperfect observations are modeled by the node $\boldsymbol{o_t}$. Actions $a_t$ are represented by rectangular decision nodes and rewards $R_t$ are drawn with diamond shape nodes. A deterioration rate POMDP can be graphically modeled by adding a deterioration rate variable $\tau_t$ instead of the time-invariant parameters $\boldsymbol{\theta_{t}}$.}
	\label{FIG:pomdpgraph}
\end{figure}
POMDPs are a generalization of MDPs in which the states are perceived by the agent through imperfect observations. The POMDP becomes a 7-tuple $\langle S, A, O, T, Z, R, \gamma\rangle$. While the dynamics of the environment are the same as for an MDP, an agent collects an observation $o\in O$ in the state $s'\in S$ with emission probability $Z(o,s',a)=P(o|s',a)$, after an action $a\in A$ is taken. Figure \ref{FIG:pomdpgraph} shows the dynamic decision network of a POMDP, which is built based on a parametric model. A deterioration rate POMDP can be equally represented if one replaces the time-invariant parameters $\boldsymbol{\theta}$ by a deterioration rate variable $\tau$. 

Since an agent is uncertain about the current state, the decisions should in principle be planned based on the full history of observations $o_1:o_t$, up to the current decision step $t$. Instead, a belief state $b(s)$ is tracked to plan the decisions. A belief state $b(s)$ is a probability distribution over states and it is updated as a function of the transition $T(s',a,s)$ and collected observation $Z(o,s',a)$: 

\begin{equation} \label{Eq:beliefupd}
b'(s') \propto P(o|s',a)\sum_{s\in S}P(s'|s,a)b(s)
\end{equation}
The normalizing constant $P(o|\mathbf{b},a)$ is the probability of collecting an observation $o\in O$ given the belief state $\mathbf{b}$ and action $a\in A$.

One can see in Eq. \ref{Eq:beliefupd} that for a specific action $a\in A$, updating a belief is equivalent to the forward operation described for DBNs in Eqs. \ref{Eq:forwOp1}-\ref{Eq:forwOpDR2}. Yet, the main objective of a POMDP is to identify the optimal policy $\pi^*(\mathbf{b})$ as a function of the belief state $\mathbf{b}$. Since the belief state is a sufficient statistic equivalent to the history of all taken actions and gathered observations, a policy $\pi^*(\mathbf{b})$ as function of $\mathbf{b}$ will always be optimal, as compared to a policy $\pi(h)$ constrained by a limited set of heuristic rules $h_{a,e}$. This is also demonstrated through numerical experiments in Section \ref{Sec:Example}. In Section \ref{Sec:POMDPImplem}, POMDP implementation details are provided and in Section \ref{Sec:PointSolvers}, we explain how point-based solvers are able to solve high-dimensional state space POMDPs and find the optimal strategies. 

\subsection{POMDP model implementation}  \label{Sec:POMDPImplem}
A systematic scheme for building a POMDP model in the context of optimal inspection and maintenance planning is provided in this section. A POMDP is built by defining all the elements of the tuple $\langle S, A, O, T, Z, R, \gamma\rangle$. While most of the reported applications of POMDPs for infrastructure planning employed a deterioration rate model \citep{Papakonstantinou2014Part2}, a parametric model as presented in Section \ref{Subsec:DBNparam} is originally implemented here.

\subsection*{States}
For the typical discrete state MDP/POMDP cases, a discretization should be first performed for continuous random variables, transforming them to the discrete state space. As mentioned in Section \ref{Sec:DBN}, an efficient discretization has to balance model fidelity and computational complexity. 

To construct an infinite horizon POMDP equivalent to the DBN parametric model presented in Section \ref{Subsec:DBNparam}, the states $S_t = S_{d_t} \times S_{\boldsymbol{\theta}}$ are assigned as the domain instances of the joint probability $P(d_{t},\boldsymbol{\theta})$. POMDPs are often represented in robotics applications by Markov hidden models containing only one hidden random variable. This has induced some confusion in the literature, where it is reported that POMDPs cannot handle deterioration mechanisms as function of time-invariant parameters \citep{JanniePOMDP2015}. However, a deterioration represented by time-invariant parameters can be easily modeled with POMDPs by augmenting the state space to include the joint probability distribution $P(d_{t},\boldsymbol{\theta})$. While state-augmentation techniques have been already proposed in the literature \citep{Papakonstantinou2014Part2,robelin2007history, kim2018pomdp}, we particularly augment the state space here in order to specify the POMDP dynamics based on deterioration processes modeled as parametric DBNs that also include time-invariant parameters. This approach can also accommodate formulations with model updating. Naturally, augmenting the state space implies an increase of computational complexity, as is the case for both DBNs and POMDPs.

If the deterioration rate model (Section \ref{Subsec:DBNdetrat}) is instead preferred, the states $S_t = S_{d_t} \times S_{\tau_t}$ are defined directly from the domain of the joint probability $P(d_{t},\tau_t)$. The implementation for this case is documented in \citep{Papakonstantinou2014Part2}. At the initial time step, one can prescribe the initial belief $\mathbf{b_0}$ as the joint probability $P(d_{t=0},\boldsymbol{\theta})$ or $P(d_{t=0},\tau_0)$. 

\subsection*{Action-observation combinations}
Actions $a\in A$ correspond to maintenance actions, such as ``do-nothing", ``perfect-repair" or ``minor-repair", and observation action $e\in E$ are defined based on the available inspection or monitoring techniques, such as ``no-observation", ``visual-inspection" or ``Nondestructive Evaluation (NDE)-inspection". 

Since rewards are assigned as a result of an agent who takes an action and perceives an observation, it is recommended to combine actions and observations into groups \citep{Papakonstantinou2014Part2}. For instance, one can combine the action ``do-nothing" with two inspections, resulting in the two combinations: ``do-nothing / visual-inspection" or ``do-nothing / NDE-inspection" and a relevant reward will be assigned to each combination. 

\subsection*{Transition probabilities}
A transition matrix $T(s,a,s')$ models the transition probability $P(s'|s,a)$ of a component from state $s\in S$ to state $s'\in S$ after taking an action $a\in A$. Therefore, the transition matrix is constructed as a function of the maintenance actions:
\begin{itemize}
    \item Do-nothing (DN) action: there is no maintenance action planned in this case and the state evolves according to the stochastic deterioration process. For an infinite horizon POMDP, the transition probability $T(s,a_{DN},s')$ is equal to the transition matrix $P(d_{t+1},\boldsymbol{\theta_{t+1}}|d_{t},\boldsymbol{\theta_{t}})$ or  $P(d_{t+1},\tau_{t+1}|d_{t},\tau_{t})$, derived in Section \ref{Sec:DBN}.
    \item Perfect repair (PR) action: a maintenance action is performed and the component returns from its current damage belief $\mathbf{b_{t}}$, at time step $t$, to the belief $\mathbf{b_{0}}$, associated with an intact status. In a belief space environment, a perfect repair transition matrix is defined as:

\begin{equation} \label{Eq:PerRepairT}
\mathbf{P}(s'|s,a_{PR}) = 
\begin{pmatrix}
b_{0}(s_0) & b_{0}(s_1) & \cdots & b_{0}(s_k) \\
b_{0}(s_0) & b_{0}(s_1) & \cdots & b_{0}(s_k)  \\
\vdots  & \vdots  & \ddots & \vdots  \\
b_{0}(s_0) & b_{0}(s_1) & \cdots & b_{0}(s_k)  
\end{pmatrix}
\end{equation}
    
    Since the belief state is a probability distribution, the summation over all the states is equal to one ($\sum b_{t}(s) =1$). If one multiplies a belief state by the transition matrix defined in Equation \ref{Eq:PerRepairT}, the current belief returns to the belief $\mathbf{b_0}$, independently of its current condition as:
\begin{equation} \label{Eq:PerRepairT2}
{b}_{0}(s)={b}_{t}(s)\: \mathbf{P}(s'|s,a_{PR}) 
\end{equation}

    If the states are fully observable, the belief state becomes a zero-one vector and a perfect repair matrix can be formulated as $\mathbf{P}(s_0|s_t,a_{PR}) = 1$,  transferring any state $s_t$ to the intact state $s_0$.

    \item Imperfect repair (IR) action: a maintenance action is performed and the component returns from a damage belief $\mathbf{b_{t}}$ to a healthier damage state or more benign deterioration rate. The definition of the repair transition matrix $\mathbf{P}(s_{t+1}|s_t,a_{IR})$ is thus case dependent. Some examples can be found in \citep{Papakonstantinou2014Part2}.
    
\end{itemize}

\subsection*{Observation probabilities}
An observation matrix $Z(o,s',a)$ quantifies the probability $P(o|s',a)$ of perceiving an observation $o\in O$ in state $s'\in S$ after taking action $a\in A$. Note that we denote the observation action as $a$ to be coherent with usual POMDP formulation; yet the observation action could be also named as $e$ to be consistent with the nomenclature used in Section \ref{Subsec:heurules}. The relevant observation actions considered here are:   

\begin{itemize}
    \item No observation (NO): the belief state should remain unchanged after the transition as no additional information is gathered. The emission probability $\mathbf{P}(o|s',a_{NO})$ can be modeled as a uniform distribution over all observations. Alternatively, it can be modeled as $\mathbf{P}(o_0|s',a_{NO})=1$. The former is recommended as it will speed up the computation \citep{Papakonstantinou2014Part2}.
    
    \item Discrete indication (DI): the likelihood $P(o|s',a_{DI})$ is modeled as a discrete event, for instance, a binary indication: detection or no-detection. The likelihood is usually quantified for the binary case by a Probability of Detection (PoD) curve. A $PoD(s')$ is equivalent to the probability $P(o_D|s')$ of collecting an observation $o_D\in O$ as function of the state $s'\in s$, and the emission probability can be directly implemented as $P(o_D|s',a_{DI}) = PoD(s')$. The implementation can be equally applied for a higher dimensional discrete observation space. 

    \item Continuous indication (CI): the likelihood $P(o|s',a_{CI})$ is modeled as a continuous distribution, for example, a direct measure of a crack. In this case, the observation space must be discretized into a finite set of observations. 
\end{itemize}

\subsection*{Rewards}
An agent having a belief $\mathbf{b}$, receives a reward $R(\mathbf{b},a)$ after taking an action $a\in A$ and collecting an observation $o\in O$. In a MDP, the reward $R(s,a)$ is defined as a function of the state, while in a POMDP, the reward $R(s,a)$ is weighted over the belief state $\mathbf{b}$ to finally obtain $R(\mathbf{b},a)$:  
\begin{equation} \label{Eq:RewBelief}
R(\mathbf{b},a) =\sum_{s\in S} b(s) R(s,a)
\end{equation}
For ease of notation, the reward model is formulated hereafter based on the same notation used for the definition of the RBI cost model in Section \ref{Sec:RBI}. If desired, societal, environmental, and other consequences can also be incorporated to the reward model. In the context of infrastructure planning, the state cost $C(s,a,s')$ is defined depending on the action-observation combination. Some recommendations are listed below:
\begin{itemize}
    \item Do-nothing/no-observation (DN/NO): this case corresponds to computing the failure risk. Once the failure state subspace $S_F\subseteq S$ is defined, the annual failure probability is the probability $P(S_F'|S)$ of reaching any state in the failure state subspace $S'_F$ from the state space $S$. Alternatively, Equation \ref{Eq:RewFail} defines the cost $C_F(s,a_{DN-NO})$ only as a function of the initial state $s\in S$, if the transition matrix $P(s'|s,a)$ is implicitly considered. This option leads to a faster computation with a point-based solver, as explained subsequently. The cost value $\bar{C}(s,a_{DN-NO})$ is equal to the failure cost $C_f$ if $s\in S_F$, and equal to 0, otherwise:
\begin{equation} \label{Eq:RewFail}
C_F(s,a_{DN-NO})=\sum_{s'\in S_F} \left\{P(s'|s,a_{DN-NO}) \, C_f  \right\}
- \bar{C}(s,a_{DN-NO})
\end{equation}
\item Do-nothing/observation (DN/O): the cost is equal in this case to the one related failure risk plus one inspection cost. Both discrete and continuous indications can be included in this category. One can therefore compute the cost $C_O(s,a_{DN-O})$ just by further considering the inspection cost $C_i$:
\begin{equation} \label{Eq:RewDN0}
C_O(s,a_{DN-O})=C_F(s,a_{DN-NO})+C_i
\end{equation}
\item Repair/no-observation (R/NO): the cost $C_R (s,a_{R-NO})$ is equal to the repair cost $C_r$: 
\begin{equation} \label{Eq:RewRep}
C_R(s,a_{R-NO})=C_r
\end{equation}
The cost $C_R(s,a_{R-O})$ for a repair/inspection combination can be similarly defined by including also the inspection cost $C_i$ along with the repair cost $C_R(s,a_{R-NO})$. 
\end{itemize}

\subsection{Point-based POMDP solvers}  \label{Sec:PointSolvers}
In principle, one could apply a value iteration algorithm \citep{Kaelbling_POMDP_1998} to solve a POMDP. While value updates are computed in a $|S|$-dimensional discrete space for an MDP, value updates for POMDPs should be instead computed in a $(|S|-1)$-dimensional continuous space. The computation thus scales up considerably with the number of dimensions, increasing the computational complexity. This fact is denoted as the curse of dimensionality. Moreover, planning in a horizon $t_N$ also suffers from the curse of history, as the number of potential action-observation histories scales exponentially with the number of time steps. Hence, solving POMDPs by applying a value iteration algorithm to the whole belief state space $\mathbb{B}$, or even to a discretized belief space grid, becomes computationally intractable for practical problems.

Relatively recent, however, point-based solvers have emerged able to solve high-dimensional state space POMDPs. Point-based solvers compute value updates only based on a representative set of belief points. Several point-based solvers \citep{sarsop_POMDP,FRTDP_Conf2006,PerseusCONFSpaan2005} have been presented in the literature. Their main difference is their basis for selecting the set of representative belief points. The reader is directed to \citep{Kostas_MOMDP_POMDP} for a detailed analysis of point-based solvers applied to infrastructure planning problems. 

In an I\&M planning context, the main objective is to identify the optimal policy, as explained in Section \ref{Sec:RBI}. Instead of constraining the policy space with pre-defined decision rules, POMDPs' main objective is to find the sequence of actions ${a_0,...,a_t}$ that maximizes the expected sum of rewards for each belief $\mathbf{b} \in \mathbb{B}$. The value function is then formulated as a function of beliefs: 
\begin{equation} \label{Eq:ValueFuncPOMDP}
V^{*}(\mathbf{b})=\max_{a\in A}\left[\,\sum_{s \in S}b(s)R(s,a)+\gamma \sum_{o \in O}P(o|\mathbf{b},a)V^{*}(\mathbf{b_{s'}})\right]
\end{equation}
It is demonstrated in \citep{Piineau_POMDP_2003} that the value function is piece-wise linear and convex when it is solved exactly. The piece-wise linearity property is related to an effective value function parametrization by a set of hyper-planes or $\alpha$-vectors $\in \Gamma$, each of them associated with an action $a \in A$.  The optimal policy $\pi^*(b)$ can be selected by identifying the $\alpha$-vectors that maximize the value function $V^{*}(\mathbf{b})$:
\begin{equation} \label{Eq:ValueFuncalpha}
V^{*}(\mathbf{b})=\max_{\alpha\in \Gamma} \sum_{s\in S}\alpha(s)b(s)
\end{equation}
The convexity property now is associated with the value of information theory \citep{howard1966information}, i.e. lower-entropy states result in better decisions and as such have higher expected values than higher-entropy states. Both of these properties of piece-wise linearity and convexity can be easily visualized in up to 4D state spaces, e.g. in \citep{Papakonstantinou2014Part1}. Naturally, in applications where the state space is augmented, as explained in Section \ref{Sec:POMDPImplem}, the belief still remains a probability over states and the value function preserves its piece-wise linearity and convexity at this newly defined, enhanced state space.
\subsection*{Finite horizon POMDPs} 
Existing point-based solvers are mostly able to solve large state space problems for infinite horizon POMDPs \citep{FinitehorizonPOMDP2019}. However, an infinite horizon POMDP can be transformed to a finite horizon one by augmenting the state space, as proposed by \citep{andriotis2019value,Papakonstantinou2014Part1,Papakonstantinou2014Part2}. In this case, the time must be encoded in the state space and a terminal state is required. Note that the resulting transition, observation and reward matrices will be very sparse. Yet, it remains essential to augment the space efficiently by taking into consideration the nature of the decision-making problem. Some recommendations are listed below:

\begin{itemize}
    \item Parametric model: the transition model is stationary. Then, the same transition matrix built for an infinite horizon POMDP can be reused for any time step of the augmented, finite horizon POMDP. To ensure a finite horizon, the last time step must include an absorbing state. An infinite horizon POMDP with $|S|$ states and $|A|$ actions can be augmented to a $|A|\:|S|\:t_N+|S|+1$ finite horizon one with horizon $t_N$. 
    \item Deterioration rate model: the state space can be efficiently formatted if the component experiences only one deterioration rate per time step. This way, one deterioration rate is considered at the first time step, two deterioration rates at the second time step, and so on, incorporating one additional deterioration rate per step until the last time step is reached. An absorbing state must also be included at the end. A deterioration rate model with $|S_d|$ states, spanning over a $t_N$ horizon and two actions (do-nothing and one maintenance action) becomes a finite horizon POMDP with $\{(t_N+1)^2|S_d| + (t_N+1)|S_d|\}/2+1$ states.  Additional maintenance actions can be included without an increase of the state space if they do not introduce additional/new deterioration rates. 
\end{itemize}

\section{Numerical experiments: Crack growth represented by time-invariant parameters.} \label{Sec:Example}
With the main objectives of providing implementation details for the two presented POMDP formulations, as well as quantifying the differences in policies and costs between POMDP and heuristic-based I\&M approaches, a set of numerical experiments is performed in this section. All computations are conducted on an Intel Core $i9-7900X$ processor with a clock speed of $3.30 \:GHz$. The experiments consist in identifying the optimal I\&M strategy for a structural component subjected to fatigue deterioration. In particular, the first presented I\&M planning setting (in Section \ref{Subsec:Case1}) is inspired by an earlier investigation of risk-based maintenance planning methods \citep{JanniePOMDP2015}. In that study, the fatigue deterioration model was approximated by a 2-parameter Weibull distribution, whereas a physically-based crack growth model is directly utilized here. According to this fracture mechanics model, the crack size $d_{t+1}$ is computed as a function of the crack size at the previous time step $d_t$: 
\begin{equation} \label{Eq:ExamCrackGrow}
d_{t+1} =\bigg[ \Big(1-\frac{m}{2}\Big) C_{FM}S_{R}^m\pi ^{m/2}n + d_t^{1-m/2}\bigg] ^{2/(2-m)}
\end{equation}
\begin{figure}
    \begin{subfigure}{0.50\textwidth}
    		\includegraphics[scale=1]{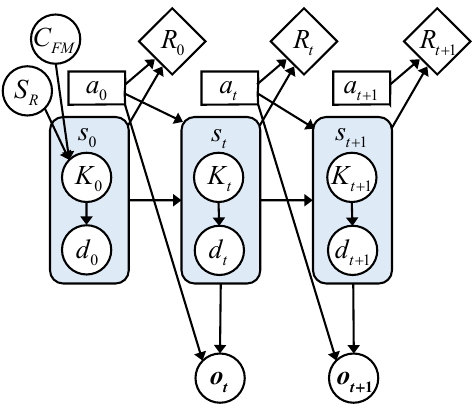}
    	\caption{Parametric POMDP }
    	\label{FIG:bpomdpSAR1}
    \end{subfigure}
    \hfill
    \begin{subfigure}{0.50\textwidth}
         \centering
    		\includegraphics[scale=1]{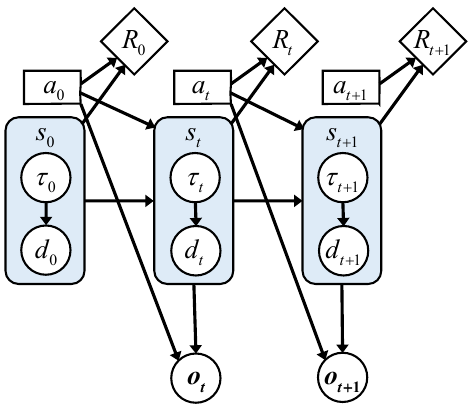}
    	\caption{Deterioration rate POMDP}
    	\label{FIG:bpomdpSAR2}
    \end{subfigure}
    \caption{Graphical representation of the POMDPs utilized for the numerical experiments. A parametric POMDP and a deterioration rate POMDP are created from the DBNs displayed in Figure \ref{FIG:dbngraph} and Figure \ref{FIG:dbngraphDR}, respectively. Note that the random variables $C_{FM}$ and $S_{R}$ are combined into the variable $K$.}
    \label{FIG:pomdpexam1}
\end{figure}
This Markovian model is derived from Paris' law, as shown in \citep{Ditlevsen2007StructuralMethods}. The process uncertainty is incorporated through the random variables listed in Table \ref{Tab:exam1par}, where $S_{R}$ stands for stress range, $C_{FM}$ corresponds to a crack growth parameter, and $d_0$ represents the initial crack size. While the crack distribution evolves over time, the parameters $C_{FM}$ and $S_{R}$ are time-invariant random variables. The remaining parameters, i.e. the crack growth parameter $m$ and the number of cycles $n$ are considered deterministic. The component fails once the crack exceeds the plate thickness $d_c$ and its considered life spans over a finite horizon $t_N$ of 30 years.

\begin{table}[h!]
\caption{Random variables and deterministic parameters utilized to model fatigue deterioration.}\label{Tab:exam1par}
\begin{tabular}{llll}
\toprule
Variable & Distribution & Mean & Standard Deviation\\
\midrule
$ln(C_{FM})$ & Normal & $-35.2$ & $0.5$ \\
$S_{R}(N/mm^{2})$ & Normal & $70$ & $10$ \\
$d_0(mm)$ & Exponential & $1$ & $1$ \\
$m$ & Deterministic & $3.5$ & - \\
$n(cycles)$ & Deterministic & $10^6$ & - \\
$t_N(yr)$ & Deterministic & $30$ & - \\
$d_c(mm)$ & Deterministic & $20$ & - \\
\bottomrule
\end{tabular}
\end{table}

\subsection{Discretization analysis}  \label{Subsec:Discret}
\begin{table}
\caption{Description of the discretization schemes considered in the sensitivity analysis, for both parametric and deterioration rate POMDP models.}\label{Tab:exam1discint}
\begin{tabular}{cl}
\toprule
Variable & Interval boundaries\\
\midrule
\multicolumn{2}{l}{\textbf{Parametric model}} \\
$S_d$ & $0, \mathrm{exp}\Big\{ \mathrm{ln}(10^{-1}):\dfrac{\mathrm{ln}(d_{c}) - \mathrm{ln}(10^{-1})}{|S_d|-2}:\mathrm{ln}(d_{c})\Big\},\infty $ \\
$S_K$ & $0, \mathrm{exp}\Big\{ \mathrm{ln}(10^{-5}):\dfrac{\mathrm{ln}(1)-\mathrm{ln}(10^{-5})}{|S_K|-2}:\mathrm{ln}(1)\Big\},\infty $ \\
\midrule
\multicolumn{2}{l}{\textbf{Deterioration rate model}} \\
$S_d$ & $0, \mathrm{exp}\Big\{ \mathrm{ln}(10^{-4}):\dfrac{\mathrm{ln}(d_{c})-\mathrm{ln}(10^{-4})}{|S_d|-2}:\mathrm{ln}(d_{c})\Big\},\infty $ \\
$S_{\tau}$ & $0:1:30$ \\
\bottomrule
\end{tabular}
\end{table}

A discretization analysis is performed to select an appropriate state space for this application. As explained in Section \ref{Sec:DBN}, either a parametric model or a deterioration rate model can be used to track the deterioration. The transition models are calculated, for both DBN models, based on data collected from simulations of the fracture mechanics model in Equation \ref{Eq:ExamCrackGrow}. The POMDPs associated with these models are graphically represented in Figure \ref{FIG:pomdpexam1}. Note that the parameters $C_{FM}$ and $S_{R}$ are grouped together for the parametric model, resulting in a new parameter $K$. By combining two random variables into one, we alleviate computational efforts \citep{StraubDBN2009}. $K$ thus corresponds to $C_{FM}S_{R}^m\pi ^{m/2}n$.

The main purpose of a proper discretization is to allocate the relevant intervals so that a high accuracy is achieved, without hindering computational tractability. Although several simulations were run, the reported results are mainly related to the case in which two inspections are planned at years 18 and 25, resulting in a no-detection outcome. The inspection quality is quantified with a probability of detection curve $PoD (d) \sim Exp[\mu = 8]$. 
A crude Monte Carlo Simulation (MCS), containing $10^7$ samples, was run to estimate the cumulative failure probability $P_{F_{MCS}}$ (Equation \ref{Eq:Upd1}). The accuracy is quantified here as the squared difference between $P_{F_{MCS}}$ and the cumulative failure probability $P_F$ retrieved by each discretized state space model. $P_F$ was obtained by unrolling a DBN over time. Note that $P_F$ can be calculated directly through a DBN, as the probability of being in the failure states of $d$. Both $P_{F_{MCS}}$ and $P_F$ are normalized to $\bar{P}_F = (P_{F} - \mu_{P_{F-MCS}}) / \sigma_{P_{F-MCS}}$, where $\mu_{P_{F-MCS}}$ and $\sigma_{P_{F-MCS}}$ are the mean and standard deviation of the failure probabilities computed by MCS, respectively. The error $\xi$ is computed as the squared difference of $\bar{P}_{F_{MCS}}$ and $\bar{P}_{F}$ for each time step:

\begin{equation} \label{Eq:ExamLSQDiff}
\xi = \sum_{t=0}^{N} \Big[ \bar{P}_{F_{MCS}}(t) - \bar{P}_{F}(t)\Big]^2
\end{equation}

\begin{figure}[h!]
		\includegraphics[scale=1.0]{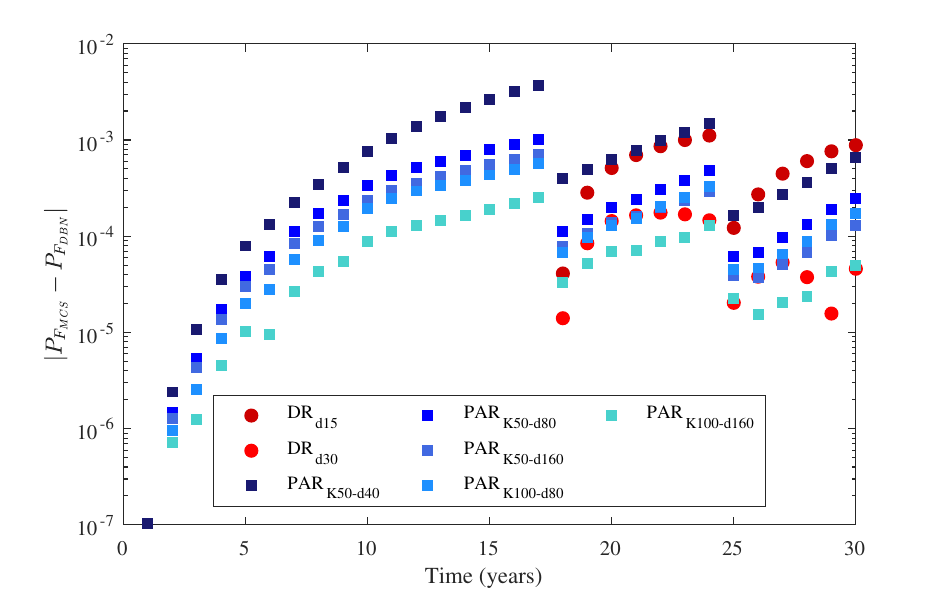}
	\caption{Error $|P_{F_{MCS}} - P_{F_{DBN}}|$ between the continuous deterioration model and the considered discrete space models. The continuous model is computed by a Monte Carlo simulation of 10 million samples and is compared with discrete state-space DBN models. The circles in the graph represent the error from deterioration rate models and the squares represent the error from parametric models.}
	\label{FIG:examdiscr}
\end{figure}
\begin{table}[h!]
%\centering
\caption{Accuracy of the considered discretization schemes. The normalized error $\xi$ and state spaces corresponding to each parameter are reported.}\label{Tab:exam1errorDisc}
\begin{tabular}{llllll}
\toprule
Model & $|S_K|$ & $|S_{\tau}|$ & $|S_d|$ & $|S|$ & $\xi$ \\
\midrule
Deterioration rate ($DR_{d15}$) & - & 31 & 15 & 465 & $8.6\cdot10^{-3}$ \\
Deterioration rate ($DR_{d30}$) & - & 31 & 30 & 930 & $2.1\cdot10^{-4}$ \\
Parametric ($PAR_{K50-d40}$) & 50 & - & 40 & 2,000 & $7.1\cdot10^{-2}$ \\
Parametric ($PAR_{K50-d80}$) & 50 & - & 80 & 4,000 & $7.2\cdot10^{-3}$ \\
Parametric ($PAR_{K50-d160}$) & 50 & - & 160 & 8,000 & $3.4\cdot10^{-3}$ \\
Parametric ($PAR_{K100-d80}$) & 100 & - & 80 & 8,000 & $2.5\cdot10^{-3}$ \\
Parametric ($PAR_{K100-d160}$) & 100 & -  & 160 & 16,000 & $4.3\cdot10^{-4}$ \\
\bottomrule
\end{tabular}
\end{table}
Table \ref{Tab:exam1discint} lists the discretization intervals for both parametric and deterioration rate models. Since the discretization is arbitrary, the interval boundaries were selected by trial and error, according to the recommendations proposed in \citep{StraubDBN2009}, i.e. a logarithmic transformation is applied to both $S_d$ and $S_k$ spaces.  Different state spaces were also tested by varying the number of states for $|K|$ and $|d|$. Table \ref{Tab:exam1errorDisc} reports the error $\xi$ for each considered state space. While the deterioration rate model of 930 overall states results in an error of magnitude less than $10^{-3}$, the state space of the parametric model is increased up to 16,000 overall states to achieve an error of magnitude less than $10^{-3}$. To illustrate the differences between the analyzed models, Figure \ref{FIG:examdiscr} shows the unnormalized error $|P_{F_{MCS}} - P_{F_{DBN}}|$ for each case. The error of the deterioration rate model is negligible before the first inspection update at 18 years, while the parametric model accumulates error throughout the whole analysis. 

\begin{table}[h!]
\caption{Analytical (AN) and simulation-based (SIM) comparison between POMDPs and optimized heuristic-based policies in a traditional setting. $\mathbf{E}[C_T]$ is the total expected cost and $\Delta \%$[POMDP FH] indicates the relative difference between each method and SARSOP finite horizon POMDP. Confidence intervals on the expected costs, assuming Gaussian estimators, are listed for the simulation-based cases.}\label{Tab:exam1theorresult}
\begin{threeparttable}
\begin{tabular}{lll}
\toprule
\textbf{Traditional setting} & $\mathbf{E}[C_T]$ $(95\% C.I)$ &  $\Delta \%$[POMDP FH]\\
\midrule
$\mathbf{Experiment \: R_{R/I}20-R_{F/R}100}$ \\
$C_i=5$, $C_r=10^2$, $C_f=10^4$, $\gamma=0.95$ \\
AN: POMDP Finite horizon. SARSOP - Lower bound & 58.35 & - \\
AN: Heur.\tnote{*} \hspace{0.1cm} EQ-INS $\Delta_{Ins}=4$ & 69.17 & +18\% \\
AN: Heur.\tnote{*} \hspace{0.1cm} THR-INS $\Delta P_{F_{th}}=3\cdot 10^{-4}$ & 65.62 & +12\% \\
SIM: POMDP Infinite horizon. SARSOP - 30 years\tnote{**} & 60.23 ($\pm$0.76) & +3\%\\
SIM: Heur. EQ-INS $\Delta_{Ins}=4$ & 69.02 ($\pm$0.83) & +18\% \\
SIM: Heur. THR-INS $\Delta P_{F_{th}}=3\cdot 10^{-4}$ & 64.81 ($\pm$0.75) & +11\% \\
\midrule
$\mathbf{Experiment \: R_{R/I}10-R_{F/R}10}$ \\
$C_i=1$, $C_r=10$, $C_f=10^2$, $\gamma=0.95$ \\
AN: POMDP Finite horizon. SARSOP - Lower Bound & 2.25 & - \\
AN: Heur.\tnote{*} \hspace{0.1cm} EQ-INS no inspections & 2.25 & +0\% \\
AN: Heur.\tnote{*} \hspace{0.1cm} THR-INS no inspections & 2.25 & +0\% \\
SIM: POMDP Infinite horizon. SARSOP - 30 years\tnote{**} & 2.50 ($\pm$0.02) & +11\% \\
SIM: Heur. EQ-INS no inspections & 2.25 ($\pm$0.00) & +0\% \\
SIM: Heur. THR-INS no inspections & 2.25 ($\pm$0.00) & +0\% \\
\midrule
$\mathbf{Experiment \: R_{R/I}50-R_{F/R}20}$ \\
$C_i=1$, $C_r=50$, $C_f=10^3$, $\gamma=0.95$  \\
AN: POMDP Finite horizon. SARSOP - Lower Bound & 12.45 & - \\
AN: POMDP Finite horizon. FRTDP - Lower Bound & 12.45 & +0\% \\
AN: POMDP Finite horizon. PERSEUS - Lower Bound & 12.96 & +4\% \\
AN: Heur.\tnote{*} \hspace{0.1cm} EQ-INS $\Delta_{Ins}=11$ & 17.06 & +37\% \\
AN: Heur.\tnote{*} \hspace{0.1cm} THR-INS $\Delta P_{F_{th}}=1\cdot 10^{-3}$ & 16.69 & +34\% \\
SIM: POMDP Infinite horizon (DR). SARSOP - 30 years\tnote{**}& 12.99 ($\pm$0.24) & +4\% \\
SIM: POMDP Infinite horizon (PAR). SARSOP - 30 years\tnote{**}& 13.08 ($\pm$0.23) & +5\% \\
SIM: Heur. EQ-INS $\Delta_{Ins}=11$ & 16.28 ($\pm$0.19) & +31\% \\
SIM: Heur. THR-INS $\Delta P_{F_{th}}=1.5\cdot 10^{-3}$ & 16.43 ($\pm$0.20) & +32\% \\
SIM: Heur. EQ-INS\tnote{***} \hspace{0.35cm}$\Delta_{Ins}=5$ & 14.17 ($\pm$0.26) & +14\% \\
SIM: Heur. THR-INS\tnote{***} \hspace{0.35cm}$\Delta P_{F_{th}}=8\cdot 10^{-4}$ & 13.29 ($\pm$0.23) & +7\% \\
\bottomrule
\end{tabular}
\begin{tablenotes}
            \item[*] The decision tree is simplified to one single branch, as explained in Section \ref{Subsec:heurules}.
            \item[**] Simulation of an infinite horizon POMDP policy over a horizon of 30 years.
            \item[***] Perfect repair actions are undertaken after two consecutive `detection' observations.
\end{tablenotes}
\end{threeparttable}

\end{table}   

In general, the selection of the discretized model will depend on the available computational resources and required accuracy. For this application, the deterioration rate model with 930 states is utilized for the numerical experiments, due to its reduced state space as compared to the parametric models.  
\subsection{Case 1. Traditional I\&M planning setting}  \label{Subsec:Case1}

The fatigue deterioration is modeled according to the time-invariant crack growth described at the beginning of Section \ref{Sec:Example}. In this traditional setting, the decision maker is only allowed to control the deterioration by undertaking a perfect repair and is able to collect observations through one inspection technique type. The perfect repair returns the component to its initial condition $d_0$ and the quality of the inspection technique is quantified with a $PoD (d) \sim Exp[\mu = 8]$. This I\&M decision-making problem is solved here by both POMDPs and heuristics. For the case of POMDPs, point-based solvers provide a theoretical guarantee to optimality, whereas RBI approaches can analytically compute the $\mathbf{E}[C_T]$ from a simplified decision tree, as explained in Section \ref{Sec:RBI}. Alternatively, the computation of the $\mathbf{E}[C_T]$ can be performed in a simulation environment, in which the deterioration process is modeled by DBNs and the costs are evaluated according to the predefined heuristic policies, as shown in Equation \ref{Eq:DBNSim2}. To equally compare the policies generated by POMDP and heuristics, the total expected costs $\mathbf{E}[C_T]$ are computed both on an analytical basis and in a simulation environment. 

\subsection*{Analytical comparison}

\begin{figure}[t]
		\includegraphics{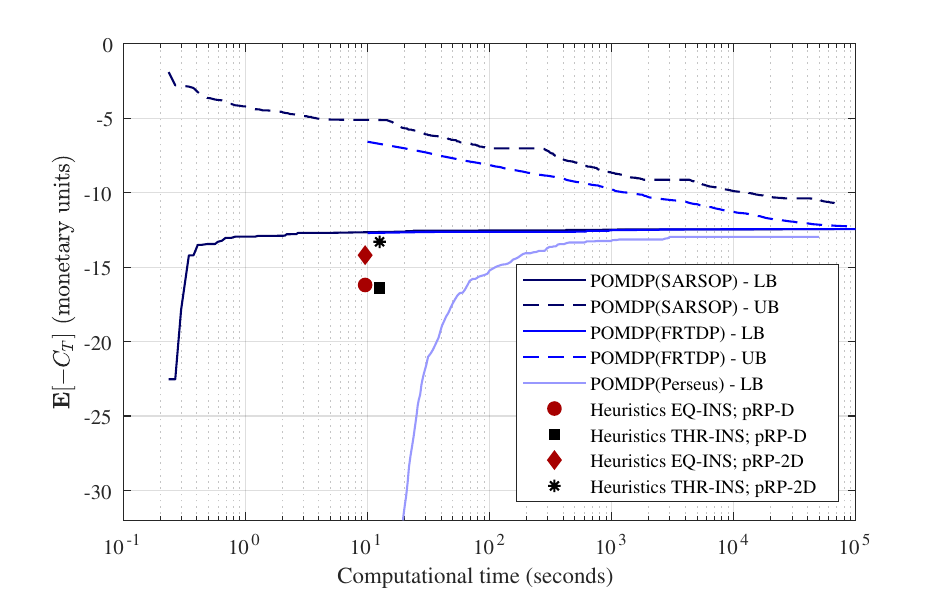}
	\caption{Point-based POMDP solutions for Experiment $R_{R/I}50-R_{F/R}20$. The expected total cost $\mathbf{E}[C_T]$ is represented over the computational time. Results of SARSOP, FRTDP and Perseus point-based POMDP solvers are plotted, with a continuous line for the low bound and a dashed line for the upper bound. Optimized heuristic methods are represented by markers; the equidistant inspection planning scheme in red, and the annual failure probability threshold in black. The markers also indicate whether the investigated heuristics plan performs repair after observing one detection outcome, $pRP-D$, or after the collection of two consecutive detection outcomes, $pRP-2D$.}
	\label{FIG:examfullpol}
\end{figure}
Following the results of the discretization analysis, a finite horizon (FH) POMDP is derived from the deterioration rate model with 930 states ($|S_d|= 30$ and $|S_{\tau}| = 31$). Since the horizon spans over 30 years, the state space is augmented from 930 to 14,880 states, as explained in Section \ref{Sec:PointSolvers}. Actions and observations are combined into three action-observation groups: (1) do-nothing/no-inspection, (2) do-nothing/inspection, and (3) perfect-repair/no-inspection. The fourth combination (repair/inspection) is not included as it will hardly be the optimal action at any time step. A total of three representative experiments are conducted, assigning different inspection, repair and failure costs to each of them. Each experiment is characterized by a different ratio between repair and inspection costs $R_{R/I}$, as well as the ratio between failure and repair costs $R_{F/R}$. Since these ratios are of relevance in this work, analyzing the problem from an optimization perspective, an explicit separation of economic, societal, and environmental consequences and their scaling to monetary units is not considered. The SARSOP point-based POMDP solver \citep{sarsop_POMDP} is used for the computation of the optimal I\&M policies. Additionally, the policies from FRTDP \citep{FRTDP_Conf2006} and Perseus \citep{PerseusCONFSpaan2005} point-based solvers are computed specifically for experiment $R_{R/I}50-R_{F/R}20$. In this theoretical comparison, the expected costs are computed based on the lower bound alpha vectors, as explained in Section \ref{Sec:PointSolvers}. 

In contrast, the optimal RBI policies are determined based on the best identified heuristic decision rules. For this theoretical comparison, the decision tree is simplified to a single branch with two schemes considered here: equidistant inspections (EQ-INS) and annual failure probability $\Delta P_F$ threshold (THR-INS). For the maintenance actions, the component is perfectly repaired after a detection indication, behaving thereafter as if a crack was not detected at that inspection. The optimized heuristics for each experiment are listed in Table \ref{Tab:exam1theorresult}, e.g. an inspection every 4 years ($\Delta_{Ins}=4$) is identified as the optimal equidistant inspection heuristic (EQ-INS) for Experiment $R_{R/I}20-R_{F/R}100$. 

The total expected cost $\mathbf{E}[C_T]$ resulting from finite horizon POMDPs and the best identified heuristics are listed in Table \ref{Tab:exam1theorresult}. Along with the $\mathbf{E}[C_T]$, the relative difference between each method and the finite horizon POMDP is also reported, and Table \ref{Tab:exam1theorresult} demonstrates that finite horizon POMDP policies outperform heuristic-based policies. Even for this traditional I\&M decision-making problem, POMDPs provide a significant cost reduction ranging from 11\% in Experiment $R_{R/I}20-R_{F/R}100$ to 37\% reduction in Experiment $R_{R/I}50-R_{F/R}20$. Experiment $R_{R/I}10-R_{F/R}10$ is merely conducted to validate the comparative results by checking that all the methods provide the same results for the case in which repairs and inspections are very expensive relatively to the failure cost.

As pointed out in Section \ref{Sec:PointSolvers}, point-based solvers are able to rapidly solve large state-space POMDPs. This is demonstrated in Figure \ref{FIG:examfullpol}, where SARSOP outperforms heuristic-based schemes in less than one second of computational time. Note that POMDP policies are based on the lower bound, whereas the upper bound, when provided, is just an approximation, to optimally sample reachable belief points \citep{Kostas_MOMDP_POMDP}. 
\begin{figure}[h!]
    \begin{subfigure}{0.50\textwidth}
    		\includegraphics[scale=1]{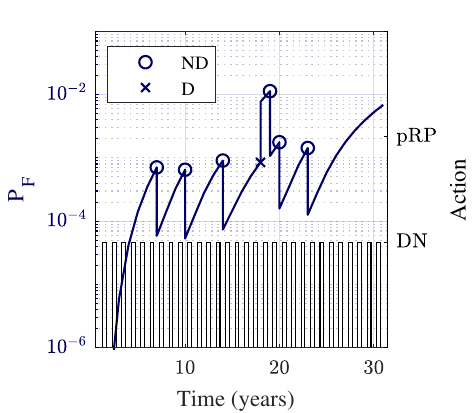}
    	\caption{POMDP policy realization}
    	\label{FIG:bpolsimSAR1}
    \end{subfigure}
    \hfill
    \begin{subfigure}{0.50\textwidth}
         \centering
    		\includegraphics[scale=1]{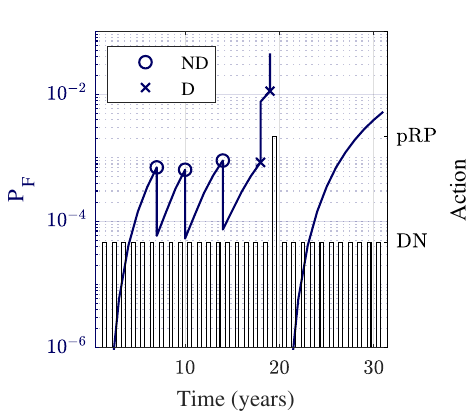}
    	\caption{POMDP policy realization}
    	\label{FIG:bpolsimSAR2}
    \end{subfigure}
    \begin{subfigure}{0.50\textwidth}
    		\includegraphics[scale=1]{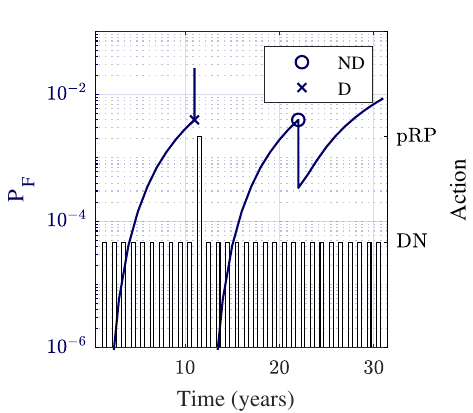}
    	\caption{Equidistant inspection policy realization}
    	\label{FIG:bpolsimInt}
    \end{subfigure}
    \hfill
    \begin{subfigure}{0.50\textwidth}
         \centering
    		\includegraphics[scale=1]{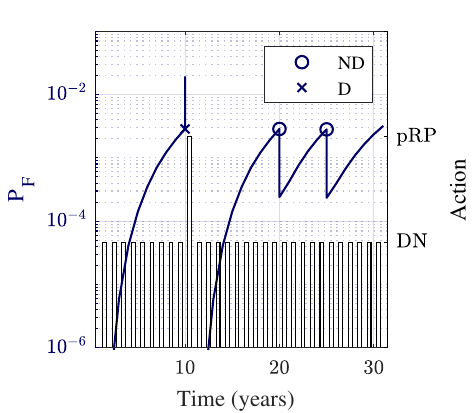}
    	\caption{$\Delta P_F$ threshold policy realization}
    	\label{FIG:bpolsimTh}
    \end{subfigure}
    \caption{Experiment $R_{R/I}50-R_{F/R}20$ policy realizations. The failure probability is plotted in blue and the prescribed maintenance actions are represented by black bars. A detection outcome is marked by a cross, whereas a no-detection outcome is marked by a circle.}
    \label{FIG:bpolsimall}
\end{figure}

\subsection*{Comparison in a simulation environment}

In this case, the total expected cost $\mathbf{E}[C_T]$ is evaluated in a simulation environment. Since the horizon can be controlled in a policy evaluation, infinite horizon POMDPs are also included in this comparison. The infinite horizon POMDP is directly derived from the deterioration rate model, and while the action-observation combinations remain the same as for the finite horizon POMDP, the belief space is now reduced to 930 states, offering a substantial reduction in computational cost, as explained before. Note that even though policies generated by infinite horizon POMDPs can be evaluated over a finite horizon, the policies are truly optimal only in an infinite horizon setting.

In this comparison, the best heuristic-based I\&M policy is also identified by analyzing two inspection planning heuristics, as previously, either based on equidistant inspections (EQ-INS) or based on an annual failure probability threshold (THR-INS). However, in this simulation setting, the component naturally returns to its initial condition after a repair, instead of modeling its evolution as a no-detection event. This operation might add a significant computational expense for analytical computations, if the decision tree is explicitly modeled; however, it can be easily modeled in a simulation-based environment. The expected utility $\mathbf{E}[C_T]$ is estimated according to Equation \ref{Eq:DBNSim2}. 

Table \ref{Tab:exam1theorresult} lists the results of the comparison and given that the expected cost $\mathbf{E}[C_T]$ is estimated through simulations, the numerical confidence bounds are also reported, assuming a Gaussian estimator. All the methods are compared relatively to the finite horizon POMDP that again outperforms the heuristic-based policies. The reduced state-space infinite horizon POMDP policy results in only a slight increment to the total expected cost obtained by the finite horizon POMDP, in this finite horizon problem. The optimal policy for an infinite horizon in experiment $R_{R/I}20-R_{F/R}100$ includes the possibility of maintenance actions, whereas the policy for a finite horizon prescribes only the action do-nothing/no-inspection. This explains the slight difference of expected costs for the infinite horizon POMDP. The infinite horizon POMDP for a parametric model of 16,000 states is also computed and listed in Table \ref{Tab:exam1theorresult} for the experiment $R_{R/I}50-R_{F/R}20$. As expected ,the $\mathbf{E}[C_T]$ for the parametric (PAR) model results in good agreement with the deterioration rate (DR) model and the small difference is attributed to the discretization quality.  

Finally, we showcase policy realizations to visualize the difference between POMDPs and heuristic-based policies over an episode, related to the experiment $R_{R/I}50-R_{F/R}20$. Figure \ref{FIG:bpolsimSAR1} and Figure \ref{FIG:bpolsimSAR2} represent realizations of POMDP policies, whereas, Figure \ref{FIG:bpolsimInt} and Figure \ref{FIG:bpolsimTh} represent the realizations of heuristic-based policies. While heuristic-based policies prescribe a repair action immediately after a detection, POMDP-based policies might also consider a second inspection after a detection outcome. If the second inspection results in a no-detection outcome, a repair action may not be prescribed; however, if the second inspection also results in detection, a perfect repair is planned. POMDP-based policies provide, therefore, more flexibility, in general, and can reveal interesting patterns, such that it might be worthy, in certain cases, to conduct a second inspection before prescribing an expensive repair action. As such, based on analyzed POMDP policy patterns, heuristic rules can be informed and defined anew. As reported in Table \ref{Tab:exam1theorresult}, two additional heuristic rules are thus examined, where perfect repair actions are undertaken after two consecutive `detection' observations. These modified heuristics yield results closer to those provided by POMDP policies, with POMDP policies surpassing now the two heuristic ones by 7\% and 14\%, respectively. While an experienced operator might have initially guessed these more sophisticated heuristic decision rules, based on the imperfect and cheap observation model specified in this setting, in more complex settings, e.g. an I\&M planning scenario with inspections that provide more than two indications (as shown in Section \ref{Subsec:Case2}), decision makers might guide their choices for the selection of more advanced heuristic rules through an investigation of the patterns exposed by POMDP policy realizations.      

\subsection{Case 2. Detailed I\&M planning setting}  \label{Subsec:Case2}
\begin{figure}
    \begin{subfigure}{0.5\textwidth}
    		\includegraphics[scale=1]{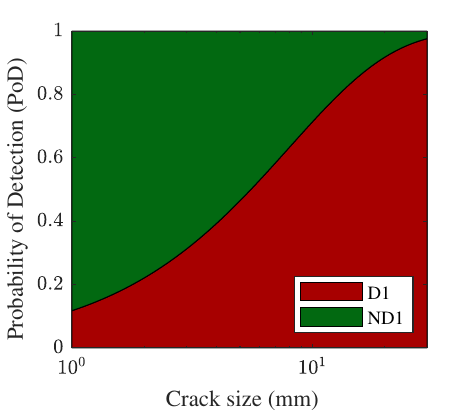}
    	\caption{Inspection type 1 (2 indicators).}
    	\label{FIG:pod1}
    \end{subfigure}
    \hfill
    \begin{subfigure}{0.5\textwidth}
         \centering
    		\includegraphics[scale=1]{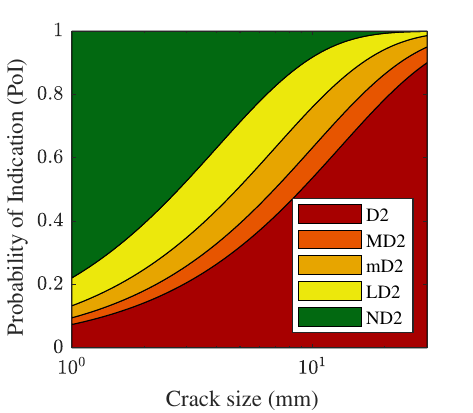}
    	\caption{Inspection type 2 (5 indicators).}
    	\label{FIG:pod2}
    \end{subfigure}
     \caption{Quantification of the inspection uncertainty. The probability of retrieving each indicator is represented as a function of the crack size. For inspection type-1 , the observation model includes two indicators: ``detection'' D1 and ``no-detection'' ND1. For inspection type-2 , the observation model is composed of five indicators: ``no-detection'' ND2, ``low damage'' LD2, ``minor damage'' mD2, ``major damage'' MD2, and ``extensive damage'' D2.}
\end{figure} 

While only a perfect repair and one inspection technique have been available for the traditional setting applications, two repair actions and two inspection techniques are now available in this more complex case. Fatigue deterioration in this setting can be controlled by either performing a perfect or a minor repair. The perfect repair returns the component to its initial condition and the minor repair transfers the component two deterioration rates back. The two inspection techniques considered are inpection 1 (I1) with only 2 indicators: detection (D) or no-detection (ND); and inspection 2 (I2) with 5 indicators: no-detection (ND), low damage (LD), minor damage (mD), major damage (MD) and extensive damage (D). The quality of each inspection technique is quantified through probability of indication (PoI) curves. Figure \ref{FIG:pod1} corresponds to the first inspection type with a $PoD (d) \sim Exp[\mu = 8]$. This inspection method is the same as the one used in the traditional I\&M planning setting. The second inspection method includes, however, the following detection boundaries: $PoI (d) \sim Exp[\mu = 4]$; $PoI (d) \sim Exp[\mu = 7]$; $PoI (d) \sim Exp[\mu = 10]$; and $PoI (d) \sim Exp[\mu = 13]$.  The probability of observing each indicator is represented in Figure \ref{FIG:pod2} as a function of the crack size. 
\begin{table}[h!]
\caption{Comparison between POMDP and optimized heuristic-based policies in a detailed setting. $\mathbf{E}[C_T]$ is the total expected cost and $\Delta \%$[POMDP FH] indicates the relative difference between each method and SARSOP finite horizon POMDP results. Confidence intervals on the expected costs, assuming Gaussian estimators, are also listed.}\label{Tab:exam2seq}
\begin{tabular}{lll}
\toprule
\textbf{Detailed setting} & $\mathbf{E}[C_T](95\% C.I)$ &  $\Delta \%$[POMDP FH]\\
\midrule
$C_{i_1}=1$, $C_{i_2}=2$, $C_{mRP}=10$, $C_{pRP}=50$, $C_f=10^3$, $\gamma=0.95$\\
POMDP Finite Horizon (FH). SARSOP - Lower Bound & 12.26  & - \\
POMDP Finite Horizon (FH). FRTDP - Lower Bound & 12.30 & $<$1\% \\
Heur. EQ-INS1 $\Delta_{Ins}=11$; $pRP$-$D1$ & 16.23 ($\pm$0.19) & +32\% \\
Heur. EQ-INS2 $\Delta_{Ins}=11$; $pRP$-$D2$ & 18.08 ($\pm$0.31) & +47\% \\
Heur. THR-INS1 $\Delta P_{F_{th}}=1.5\cdot 10^{-3}$; $pRP$-$D1$ & 16.40 ($\pm$0.20) & +33\% \\
Heur. THR-INS2 $\Delta P_{F_{th}}=1.1\cdot 10^{-3}$; $pRP$-$D2$ & 15.55 ($\pm$0.21) & +26\% \\
Heur. THR-INS2 $\Delta P_{F_{th}}=5.0\cdot 10^{-4}$; $pRP$-$P_{F_{th}}=2.2\cdot 10^{-2}$ & 13.88 ($\pm$0.29) & +13\% \\
Heur. THR-INS2 $\Delta P_{F_{th}}=1.0\cdot 10^{-3}$; $pRP$-$\mathbf{E}[d]>4$ & 13.66 ($\pm$0.24) & +11\% \\

\bottomrule
\end{tabular}
\end{table}
\begin{figure}[h!]
		\includegraphics{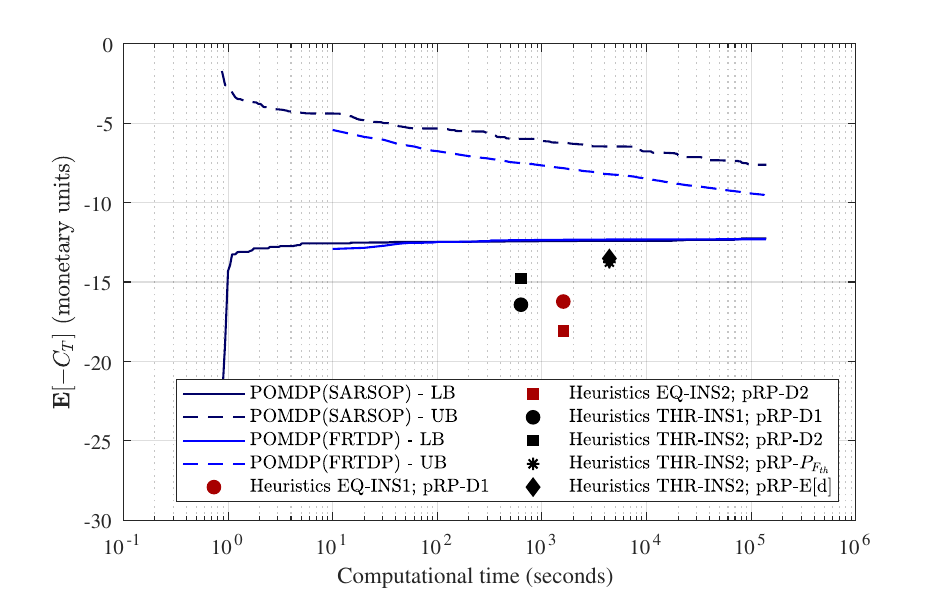}
	\caption{Computational details of POMDP and simulation-based heuristic schemes in a detailed setting. The expected total costs $\mathbf{E}[C_T]$ are represented over the computational time. Results of SARSOP and FRTDP point-based POMDP solvers are plotted, with a continuous line for the low bound and a dashed line for the upper bound. Optimized heuristic policies results are reported by markers and are directly linked to the schemes shown in Table \ref{Tab:exam2seq}.}
	\label{FIG:examCCpuTime}
\end{figure}

Similar to the previous case, we solve a finite horizon POMDP with 14,880 states to identify the optimal policy. However, in this setting, actions and observations are combined into seven groups: (1) do-nothing/no-inspection (DN-NI); (2) do-nothing/inspection-1 (DN-I1); (3) do-nothing/inspection-2 (DN-I2); (4) minor-repair/no-inspection (mRP-NI); (5) minor-repair/inspection-1 (mRP-I1); (6) minor-repair/inspection-2 (mRP-I2); and (7) perfect-repair / no-inspection (pRP-NI), and analyses are conducted for a modified version of experiment $R_{R/I}50-R_{F/R}20$. The individual costs for this example are listed in Table \ref{Tab:exam2seq}. Inspection type-2 costs twice the cost of inspection type-1, as it is more accurate and provides more information about the deterioration. 

For this setting, heuristic inspection decision rules are prescribed considering again both equidistant inspections and annual failure probability $\Delta P_F$ threshold schemes. All heuristics are evaluated in a simulation environment, computing the expected cost $\mathbf{E}[C_T]$, as indicated in Equation \ref{Eq:DBNSim2}. Maintenance heuristic rules are accordingly defined considering the following two schemes:

\begin{itemize}
    \item Observation-based maintenance rules: a maintenance action is undertaken after an observation. For example, a minor repair is undertaken if a minor damage is observed. The number of potential observation-based maintenance rules scales to $|A_R|^{|O|}$ pairs, where, $|O|$ and $|A_R|$ are the number of observations and maintenance actions, respectively. If we consider inspection type-2, the heuristic rules result in $3^5$ combinations. Such combinatoric heuristic rules, together with failure probability thresholds or intervals for inspections, have been evaluated against POMDPs in \citep{andriotis2020constraint}. Due to the large computational cost of evaluating all possible decision rules, we evaluated only a subset of these combinations here. The most competitive set of heuristic rules for this case are listed in Table \ref{Tab:exam2seq}, e.g. the optimized equidistant inspection type-1 heuristic (EQ-INS1) prescribes an inspection every 11 years ($\Delta_{Ins}=11$), and a perfect repair after a detection observation ($pRP$-$D1$).
    \item Threshold-based maintenance rules: a maintenance action is undertaken when a specific threshold is reached after an observation. The threshold can be prescribed in terms of failure probability $P_F$ or expected damage size, as proposed in \citep{Jannie2018Computational}. We consider both cases here, i.e. a failure probability threshold $P_{F_{th}}$ and an expected damage size threshold, $\mathbf{E}[d]$. Threshold-based maintenance rules based on expected damage have also been evaluated against POMDPs in \citep{Andriotis2019ManagingLearning}.
\end{itemize}
\begin{figure}
		\includegraphics{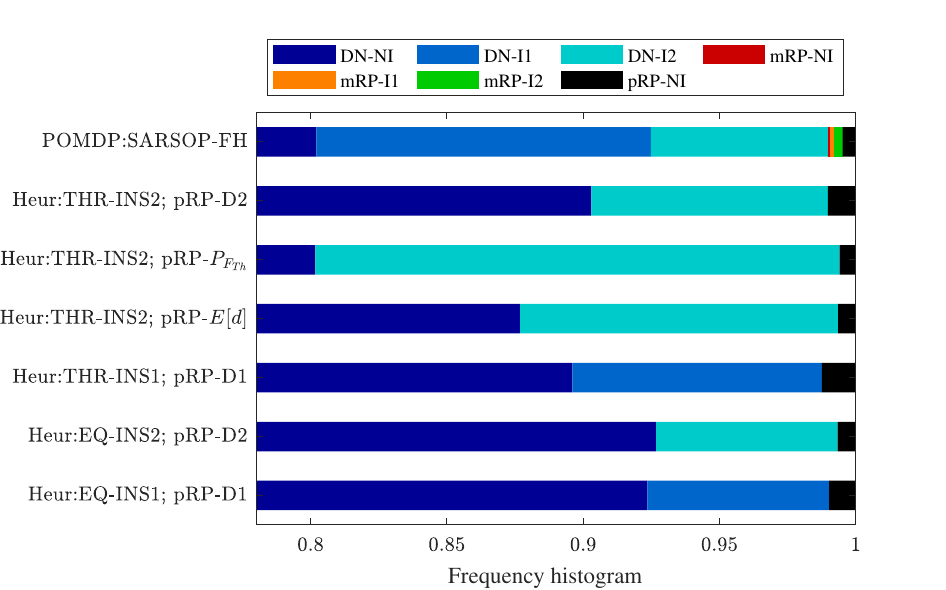}
	\caption{Frequency histogram of the actions prescribed by each considered approach over $10^4$ policy realizations. The policies presented here are linked to those listed in Table \ref{Tab:exam2seq}.}  
	\label{FIG:freqHist}
\end{figure}
The expected costs $\mathbf{E}[C_T]$ resulting from both POMDP and heuristic-based policies are reported in Table \ref{Tab:exam2seq}. Additionally, we list the relative difference between each policy and a finite horizon POMDP policy solved by SARSOP. In this detailed setting, POMDP-based policies outperform again heuristic-based ones. In terms of POMDP-based policies, SARSOP and FRTDP achieve similar results. Results obtained from heuristic-based policies vary depending on their prescribed set of heuristics. For equidistant inspection planning, inspection type-1 is preferred rather than inspection type-2, because the inspections are fixed in time, and the additional information provided by inspection type-2 becomes too expensive.
In contrast, inspection type-2 is the best scheme for annual failure probability threshold inspection planning. The threshold-based maintenance heuristics proved to be better than observation-based heuristics, yet threshold-based maintenance heuristics imply additional computational costs, as generally, more heuristic rules must be evaluated. Figure \ref{FIG:examCCpuTime} illustrates the expected cost $\mathbf{E}[C_T]$ of each policy as a function of the computational time. We can see how the POMDP point-based solvers improve their low bounds in time, along with the computational cost incurred by evaluating the various heuristic rules. 	

To visualize the actions prescribed by each approach, Figure \ref{FIG:freqHist} displays a frequency histogram of the actions taken over $10^4$ policy realizations. The action do-nothing/no-inspecion (DN-NI) predominates over all other actions. While heuristic policies conduct either inspection type-1 (DN-I1) or inspection type-2 (DN-I2), the POMDP-based policy utilizes both inspection types. This is also true for the maintenance actions, in which heuristic policies prescribe only perfect repairs, whereas POMDP policies choose sometimes to undertake minor-repairs (mRP) as well.

\section{Discussion}  \label{Sec:Discussion}
The results of this investigation show that POMDPs are able to identify optimal I\&M policies for deteriorating structures and offer substantially lower costs than heuristic-based policies, as is theoretically explained and justified, and as it has also been demonstrated through numerical examples in Sections \ref{Subsec:Case1} and \ref{Subsec:Case2}. The policy optimization based on heuristic-based approaches may be constrained by the limited number of decision rules assessed, out of all possible decision rules. Avoiding these limitations, POMDPs prescribe actions as a function of the belief state, which is a sufficient statistic of the whole, dynamically updated, action-observation history. This implies that the actions are taken according to the whole history of actions and observations, rather than as a result of an immediate inspection outcome or pre-defined static policies. 

As demonstrated in Section \ref{Subsec:Case2}, POMDPs can be applied to detailed I\&M decision settings, in which multiple actions and inspection methods are available. In terms of computational efficiency, state-of-the-art point-based solvers are able to solve high-dimensional state space POMDPs within a reasonable computational time. In particular, SARSOP point-based solver very quickly improves its policy at the beginning of the solution process and employs an approximate upper bound to gradually reach a converged solution. For both traditional and detailed settings, both SARSOP and FRTDP point-based solvers outperform heuristic-based policies after only few seconds of computational time. 

For modeling the deterioration process, one can utilize either a parametric or a deterioration rate model, as explained in Section \ref{Tab:exam1discint}. A deterioration rate model generally results in a smaller state space than a parametric model, except for very long horizons. In this latter case, a parametric model might lead to a smaller state space, due to its stationary nature. In any case, a discretization analysis must be conducted to select the appropriate state model for the problem at hand. More efforts are worth being made in the future towards continuous state space POMDPs and optimal discretization schemes for discrete state spaces.

\section{Concluding remarks}  \label{Sec:Conclusions}
In this paper, we examine the effectiveness of Partially Observable Markov Decision Processes (POMDPs) to identify optimal Inspection and Maintenance (I\&M) strategies for deteriorating structures, and we clarify that Dynamic Bayesian Networks (DBNs) can be combined with POMDPs, providing a joint framework for efficient inspection and maintenance planning. The formulation for deriving POMDPs in a structural reliability context is also presented, and two alternative DBN formulations for deterioration modeling are described, together with their POMDP implementations.

Modern Risk Based Inspection (RBI) planning methodologies are often supported by DBNs, and a pre-defined set of decision rules is evaluated. These policies can on occasions diverge significantly from globally optimal solutions, because of the limited domain space of searched policies that may not include the global optimum. In contrast, POMDP policies prescribe an action as a function of the belief state, which is a sufficient statistic of the whole action-observation history. 

I\&M policies generated by finite horizon POMDPs are compared with heuristic-based policies, for the case of a structural component subjected to fatigue deterioration. In the first example, the stochastic deterioration is modeled as a function of time-invariant parameters, with only one inspection type and one perfect repair available. Our numerical findings verify that POMDP-based policies can approximate the global solution better than heuristic-based policies, thus being more efficient even for typical RBI applications. The 14,880 states finite-horizon POMDP outperforms heuristic-based policies in less than a second of computational time.
For the second numerical example, we consider an I\&M decision-making problem in a more detailed setting, including two inspection methods and two repair actions. Whereas the outcome of the first inspection type is set up as a binary indicator, the second inspection technique indicates the damage level through five alarms. With this application, we demonstrate the capabilities of POMDPs in efficiently handling complex decision problems, outperforming again heuristic-based polices. 

The main limitation of the presented approaches, including POMDPs, is the increase of computational complexity for very large state and action spaces, such as the ones for a system of multiple components. Dynamic Bayesian networks with large state spaces are similarly constrained by the curse of dimensionality. To overcome this limitation, we suggest further research efforts toward the development of POMDP-based Deep Reinforcement Learning (DRL) methodologies. As demonstrated in \citep{andriotis2020constraint, Andriotis2019ManagingLearning}, a multi-agent actor-critic DRL approach is able to identify optimal strategies for multi-component systems with large state, action and observation spaces. In particular, POMDP-based actor-critic DRL methods approximate the policy and the value function with neural networks, alleviating therefore the curse of dimensionality through the deep networks parametrizations, and the curse of history through the reliance on dynamic programming MDP principles, the full advantages of which may be compromised if heuristic rules are instead considered.       

\section*{Acknowledgements}
This research is funded by the National Fund for Scientific Research in Belgium F.R.I.A. - F.N.R.S. This support is gratefully acknowledged. Dr. Papakonstantinou and Dr. Andriotis would further like to acknowledge that this material is also based upon work supported by the U.S. National Science Foundation under Grant No. 1751941.



\bibliographystyle{elsarticle-num} 
\bibliography{DBNPOMDPInspecMaintPlanning}




%%\bibitem{}

%%\end{thebibliography}
\end{document}